%% file: template.tex
\RequirePackage{fix-cm}
\documentclass[twocolumn]{svjour3}          
\smartqed  
\usepackage[table]{xcolor}
\usepackage{listings}
\usepackage{graphicx}

\usepackage{algorithm}
\usepackage[noend]{algpseudocode}

\usepackage{multirow, makecell}
\usepackage{enumitem}
\usepackage{pgf}
\usepackage{pgfplots}
\usepackage{tikz}
\usepackage[linguistics]{forest}
\usepackage{amsmath}
\DeclareMathOperator*{\argmax}{arg\,max}
\DeclareMathOperator*{\argmin}{arg\,min}
\usepackage{subfig}
\usepackage{graphics}
\usepackage{balance}
\usepackage{amsmath, bm}
\usepackage{pifont}
\usepackage{booktabs}

\usepackage{todonotes}

\definecolor{codegreen}{rgb}{0,0.6,0}
\definecolor{codegray}{rgb}{0.5,0.5,0.5}
\definecolor{codepurple}{rgb}{0.58,0,0.82}
\definecolor{backcolour}{rgb}{0.95,0.95,0.92}

\definecolor{r1}{RGB}{228, 26, 28}
\definecolor{r2}{RGB}{55, 126, 184}
\definecolor{r3}{RGB}{77, 175, 74}
\definecolor{r4}{RGB}{152, 78, 163}

\definecolor{ao}{rgb}{0.0, 0.5, 0.0}

\lstdefinestyle{mystyle}{
    backgroundcolor=\color{backcolour},   
    commentstyle=\color{codegreen},
    keywordstyle=\color{magenta},
    numberstyle=\tiny\color{codegray},
    stringstyle=\color{codepurple},
    basicstyle=\ttfamily\footnotesize,
    breakatwhitespace=false,         
    breaklines=true,                 
    captionpos=b,                    
    keepspaces=true,                 
    numbers=left,                    
    numbersep=5pt,                  
    showspaces=false,                
    showstringspaces=false,
    showtabs=false,                  
    tabsize=2
}
\journalname{}
\begin{document}

\title{AutoML in Heavily Constrained Applications}


\author{Felix Neutatz         \and
        Marius Lindauer \and 
        Ziawasch Abedjan
}


\institute{Felix Neutatz \at
              TU Berlin \\
              \email{f.neutatz@tu-berlin.de} 
           \and
           Marius Lindauer \at
           Leibniz Universit\"{a}t Hannover \\
           \email{m.lindauer@ai.uni-hannover.de}
           \and
           Ziawasch Abedjan \at
           Leibniz Universit\"{a}t Hannover \\
           \email{abedjan@dbs.uni-hannover.de}           
}


\input{command}

\maketitle

\begin{abstract}
Optimizing a machine learning pipeline for a task at hand requires careful configuration of various hyperparameters, typically supported by an AutoML system that optimizes the hyperparameters for the given training dataset.
Yet, depending on the AutoML system's own second-order meta-configuration, the performance of the AutoML process can vary significantly. Current AutoML systems cannot automatically adapt their own configuration to a specific use case. Further, they cannot compile user-defined application constraints on the effectiveness and efficiency of the pipeline and its generation.
In this paper, we propose \system{}, which uses meta-learning to automatically adapt its own AutoML parameters, such as the search strategy, the validation strategy, and the search space, for a task at hand. The dynamic AutoML strategy of \system{} takes user-defined constraints into account and obtains constraint-satisfying pipelines with high predictive performance. 
\end{abstract}

\input{01_introduction}

\input{02_problem}

\input{03_system}

\input{04_evaluation}
\input{05_related_work}
\input{06_conclusion}


\bibliographystyle{spmpsci}      
\bibliography{abbreviation,library}
\balance

\end{document}

%% file: command.tex
\newcommand{\ziawasch}[1]{\textcolor{blue}{Ziawasch: #1}}
\newcommand{\za}[1]{\textcolor{red}{Ziawasch: #1}}
\newcommand{\Felix}[1]{\textcolor{purple}{Felix: #1}}
\newcommand{\Binger}[1]{\textcolor{red}{Binger: #1}}

\newcommand{\newv}[1]{{\color{black} {#1}}}

\newcommand{\newvv}[1]{{\color{black} {#1}}}

\newcommand{\system}{\textsc{Caml}}

\newcommand{\coll}[1]{{\cellcolor{red}\color{black}{#1}}}
\newcommand{\collm}[1]{{\cellcolor{orange}\color{black}{#1}}}

\newcommand{\cole}[1]{{\cellcolor{blue!70!pink}\color{black}{#1}}}
\newcommand{\colem}[1]{{\cellcolor{blue!30!white}\color{black}{#1}}}

\newcommand{\answerVRone}[1]{{\color{black} {#1}}}
\newcommand{\answerVRtwo}[1]{{\color{black} {#1}}}
\newcommand{\answerVRthree}[1]{{\color{black} {#1}}}
\newcommand{\answerVRmulti}[1]{{\color{black} {#1}}}

\newcommand{\ttodo}[1]{\todo{\color{orange}hh\color{black}{#1}}}

\newcommand{\smallestfont}{\tiny}

%% file: 01_introduction.tex
\section{Introduction}

Recently, there has been intensive research on automated machine learning~(AutoML) to facilitate the design of machine learning~(ML) pipelines~\cite{feurer2015efficient,DBLP:journals/pvldb/BoehmDEEMPRRSST16,kaoudi2017cost,sparks2017keystoneml,Kumar2017Data,xin2018helix,shang2019democratizing,boehm2019systemds,Overton,DiffML,AutoPipeline,DBLP:conf/sigmod/ShahLKY021}. Existing work entails hyperparameter optimization, neural architecture search, and the generation of end-to-end ML pipelines, consisting of data preprocessing, feature engineering, model selection, and postprocessing.


\subsection{AutoML with Constraints}
In practice, AutoML can be subject to two kinds of constraints: \textit{ML application} and \textit{Search} constraints. \textit{ML application} constraints impose restrictions, such as limits on training/inference time and ML pipeline size, or additional quality criteria, such as adversarial robustness or differential privacy, on the final ML pipeline. The ML application constraints on resource consumption are particularly relevant in systems that work with dynamic data and rely on fast response time~\cite{DBLP:conf/dac/0001TRM21,DBLP:journals/tits/MehraMNC21}. 
\textit{Search} constraints impose restrictions on the AutoML search process itself, such as limiting the search time, main memory usage, or parallelism. 

Depending on the real-world setting and its commanding constraints, users have to configure the AutoML system differently to achieve the optimal result within a limited search time budget. 
With emerging applications in the realm of edge computing and real-time analysis, further constraints need to be considered. Autonomous driving relies on real-time video analysis~\cite{Elluswamy2022} and to keep up with a sufficiently high frame rate, the model has to follow tight inference time constraints. As ML models have become successful, they have also gained traction on smaller devices, such as smartphones, requiring them to reduce their memory footprints and to predict fast. For streaming use cases, it might be important to continuously retrain to adapt to concept drift over time~\cite{DBLP:conf/edbt/DerakhshanMRM19}. For fast-changing environments, such as fraud detection for high-frequency transactions, the models are subject to demanding training time constraints. Further, streaming ML requires constraints on millisecond latency and high throughput~\cite{DBLP:conf/hotos/NishiharaMWTPSL17,DBLP:journals/pvldb/GhoshGMYA22}.
There are also concerns regarding population-based quality criteria. For example, Schelter et al.~\cite{FairPrep} showed that mean-value imputation introduces bias and should be omitted from the ML hyperparameter search space if the application is subject to fairness constraints.

AutoML systems have several \emph{AutoML parameters}, such as those defining the search space, the search strategy, e.g., different variants of Bayesian optimization and evolutionary algorithms, the validation strategy, e.g., hold-out and cross-validation, and the sampling strategy, which strongly influence the search process. 
We call an arbitrary initialization of these parameters an \emph{AutoML configuration}. The \emph{default AutoML configuration} is the initialization of each AutoML parameter with its default value and typically enables the entire search space for ML hyperparameters.

Generally, there is no single AutoML configuration that always yields a model with high predictive performance on all kinds of datasets and in particular subject to any of the aforementioned constraints. Typically, expert knowledge is required to configure and adapt an AutoML system to such settings.

\subsection{Adapting AutoML Configurations}
\begin{sloppypar}
We envision an AutoML system that automatically adapts to a user-specified \emph{ML task}, i.e., not only to the dataset but also taking into account user-defined \textit{ML application constraints} and \textit{search constraints}, to achieve the best overall anytime performance. We call this new \newvv{paradigm~\emph{constraint-driven AutoML}}, where
\newv{the data scientists and domain experts who know the constraints of the ML applications upfront, e.g. resource restrictions for IoT devices or legal restrictions, only need to specify the constraints but do not need to manually adjust the space of pipeline designs. We note that AutoML addresses two groups of users: non-domain experts seeking low- or no-code solutions, and ML~experts seeking support in their day-to-day business. We rather address the latter user group with knowledge of the task-specific constraints.} 
\end{sloppypar}



State-of-the-art AutoML systems~\cite{AutoSklearn2,AutoGluon,TPOT} do not support ML application constraints out of the box, and they do not adapt the search process to user-specified search constraints. Both adaptations are in fact non-trivial because AutoML systems have many of their own parameters, such as those defining the search space, the search strategy, and the validation strategy.
For instance, if the user specifies a search time of five minutes, the well-known AutoML system Auto-Sklearn~\cite{feurer2015efficient,AutoSklearn2} will consider the same ML hyperparameter search space as if it had a whole week, although only a very small fraction of the ML hyperparameter space can be covered. Theoretically, users could modify the AutoML system parameters to reduce the search space. Still, even for experts, it is difficult to estimate which part of the ML hyperparameter space to consider or which sample size suffices for a given task. Similar to ML-hyperparameter sensitivity to the dataset at hand, AutoML's anytime performance strongly depends on its own parameters and their optimal setting depends on the ML task. An intuitive approach would be to frame the problem as a multi-objective optimization task to explore ML pipelines across all constraint dimensions. 
However, even if we consider it a multi-objective optimization problem, it is still unclear how to select the AutoML parameters to search efficiently.

\begin{sloppypar}
We propose an efficient solution for constraint-driven AutoML by leveraging meta-learning, which so far has only been applied to a few subproblems in our setting. For instance, Auto-Sklearn2~\cite{feurer2015efficient,AutoSklearn2} leverages meta-learning to warm-start Bayesian optimization (BO). Specifically, it searches for the best set of ML hyperparameters on all datasets in a repository. For a new dataset, it compares the dataset with all datasets in the repository and applies BO with an initial portfolio of  ML hyperparameter configurations of the most-similar dataset to accelerate the search. Additionally, it learns which validation strategy and initial portfolio are beneficial for which dataset. However, Auto-Sklearn2's meta-learning approach cannot support constraints because one would need to independently train the meta-learning for each possible set of constraint settings, which is infeasible.
Further, their approach only supports predicting discrete strategy decisions using pairwise meta-modeling, i.e., a meta-model predicts the better out of two possible AutoML strategies. This approach cannot handle continuous AutoML parameters, and even covering all possible combinations of sampling strategy, validation, and search space strategy is typically infeasible.
\end{sloppypar}

Another meta-learning approach is to learn a surrogate model that learns offline whether a given ML pipeline can satisfy specified constraints. However, this approach does not adapt the AutoML parameters~\cite{mohr2021predicting}. For instance, it is not possible to adapt the validation strategy based on the specified constraints.

To remedy the aforementioned limitations and to enable all degrees of freedom in constraint-driven AutoML, we addressed three major challenges:

\begin{enumerate}
  \item \textbf{Huge meta-learning space.} The combined space of AutoML parameters, constraints, and datasets is huge. 
  We need to draw meta-training instances from this huge space to enable the meta-training.   
  To prune the ML hyperparameter space, we have to consider the trade-off between search runtime and predictive performance. If we prune too much of the ML hyperparameter space, the optimization might not find ML pipelines with high predictive performance. If we prune too little, the search might be inefficient. To estimate which AutoML configurations will be successful, it is critical to consider the dataset and user-specified constraints. 
   \item \textbf{Meta-training labels.} To predict an AutoML configuration for a given task, a meta-model has to be trained on similar tasks. Choosing the right meta-training examples and an appropriate prediction target is a problem we intend to solve.
  \item \textbf{Nondeterministic AutoML.} AutoML is a nondeterministic and stochastic process. Across multiple runs, the same AutoML configuration might lead to significantly different outcomes \newv{because both the AutoML optimizer (e.g. Bayesian optimization) and ML model training are stochastic}. So, if we naively train a meta-model on such a noisy signal, the meta-model might be inaccurate.
\end{enumerate}

\subsection{Contributions}
\begin{sloppypar}
To address these challenges, we propose a new constraint-driven AutoML system, \system{}
, which dynamically configures its AutoML parameters by taking into account the user-specified ML task (i.e., dataset and constraints).
Learning from previous \emph{AutoML runs}~(i.e., dataset, constraints, AutoML configuration), \system{} generates AutoML configurations and estimates which of them are promising for a new ML task. 
To this end, we make the following contributions:
\end{sloppypar}
\begin{enumerate}[noitemsep,topsep=0em,leftmargin=*]
    \item We propose alternating sampling as a training data generation strategy - a combination of active learning, Bayesian optimization, and meta-learning. It is parallelized and efficiently explores the huge search space of datasets, AutoML configurations, and constraints to learn a meta-model that estimates the success of AutoML configurations and accelerates the search process.
    \item To instantaneously extract the most promising AutoML configurations from the meta-model at runtime, we propose offline AutoML configuration mining that provides \system{} with a large pool of promising AutoML configurations. As the meta-model can rank 100k configurations in less than a second, this pool allows for fast AutoML configuration retrieval.
    
    \item To ensure high adaptability for a wide set of constraint settings, we implemented \system{} in a way to allow the user to configure whether or not it optimizes any ML hyperparameter. It also supports ML application constraints based on metrics, such as training/inference time, ML pipeline size, and equal opportunity~\cite{hardt2016equality} - a fairness metric.
    \item We report extensive experiments with \system{} and compare it to state-of-the-art AutoML systems. We provide our implementation, datasets, and evaluation framework in our repository~\cite{Neutatz2021}.
\end{enumerate}

\noindent\textbf{Main Findings.}
\label{subsec:main_findings}
Our study lets us draw the following conclusions:
\begin{sloppypar}
\begin{enumerate}[noitemsep,topsep=0em,leftmargin=*]
    \item \system{} does not only outperform the default AutoML configuration but also state-of-the-art systems, such as TPOT~\cite{TPOT}, AutoGluon~\cite{AutoGluon}, and Auto-Sklearn2~\cite{AutoSklearn2}, in constrained settings.
    \item \system{} outperforms hand-tailored constraint-specific AutoML solutions, such as Auto-Sklearn~2~\cite{AutoSklearn2}. Manually adapting AutoML system configurations to diverse constraints or even combinations of multiple constraints is nearly impossible due to unforeseeable side effects. Therefore, solutions, such as \system{}, are required.
    \item \system{} is the first step towards our vision of constraint-driven AutoML. This way, we can cover multiple diverse constraints and add/remove additional ones without AutoML systems expertise.
\end{enumerate}
\end{sloppypar}

%% file: 02_problem.tex
\section{Three-Step Problem}
\label{sec:problem}

\begin{figure*}
	\centering
	\includegraphics[width=\textwidth]{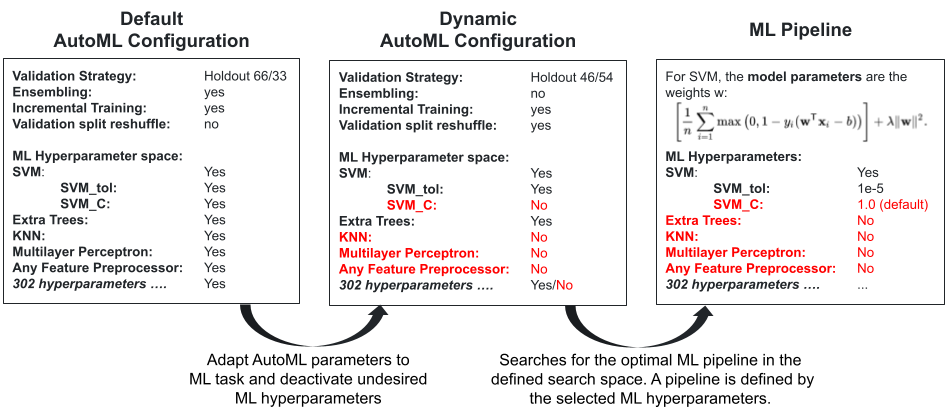}
	\caption{Example: Adapting the AutoML Parameters for constraint-driven AutoML.}
	\label{figure:parametertypes}
\end{figure*}

The three-step problem represents the search for the optimal setting of three parameters as described in Figure~\ref{figure:parametertypes}: the AutoML parameters, ML hyperparameters, and model parameters.

Before we formalize the problem of constraint-driven AutoML, we formally define the problem of finding optimal model parameters for a given supervised machine learning model and the AutoML problem of finding the optimal algorithm and ML hyperparameters, e.g., selecting a data encoding, feature preprocessor, and classification model, and all their corresponding hyperparameters.

\subsection{Supervised ML Problem}
The supervised ML problem is to find the parameters~${\bm{\theta}}$ for a predictive model~$f$ by minimizing the loss $\mathcal{L}$ of mapping $f: \bm{x}_{i} \mapsto \hat{y}_{i}$ for a given training dataset $D_{train} = \{(x_{0}, y_{0}), ..., (x_{n}, y_{n})\}$. 

\begin{equation}
\bm{\theta}^* \in  \argmin_{\bm{\theta} \in \Theta} \sum_{(x_i,y_i) \in D_{train}} \mathcal{L}_{train}\left(y_i, f(x_i; \bm{\theta}) \right).
\label{eq:mlobj}
\end{equation}

In practice, the problem is often more complex since the loss might be regularized to achieve better generalization performance, and stochastic optimizers might lead to different model parameters returned by the learning process. 

\subsection{The AutoML Problem}
The combined algorithm selection problem and hyperparameter optimization problem of AutoML~\cite{ThorntonHHL13} is to determine the predictive pipeline~$a \in A$ and its corresponding hyperparameters $\bm{\lambda} \in \Lambda$, inducing a model $f^{(a_{\bm{\lambda}}; D_{train})}(\cdot;\hat{\bm{\theta}})$ with some approximated model parameters~$\hat{\bm{\theta}}$, that achieve the lowest loss on the validation set~$D_{valid}$. Formally:

\begin{equation}
\argmin_{a \in A, \bm{\lambda} \in \Lambda} \sum_{(x_i,y_i) \in D_{val}} \mathcal{L}_{val}( y_i, f^{(a_{\bm{\lambda}}; D_{train})}(x_i; \hat{\bm{\theta}})).
\label{eq:automlobj}
\end{equation}

We note that the training loss $\mathcal{L}_{train}$ (e.g., cross-entropy) does not have to be the same as the validation loss $\mathcal{L}_{val}$ (e.g., balanced accuracy).
Since the ML model training can already take some time (e.g., training a DNN), AutoML has to be very efficient in evaluating different configurations from $A \times \Lambda$.

\begin{figure*}
	\centering
	\includegraphics[scale=1.0]{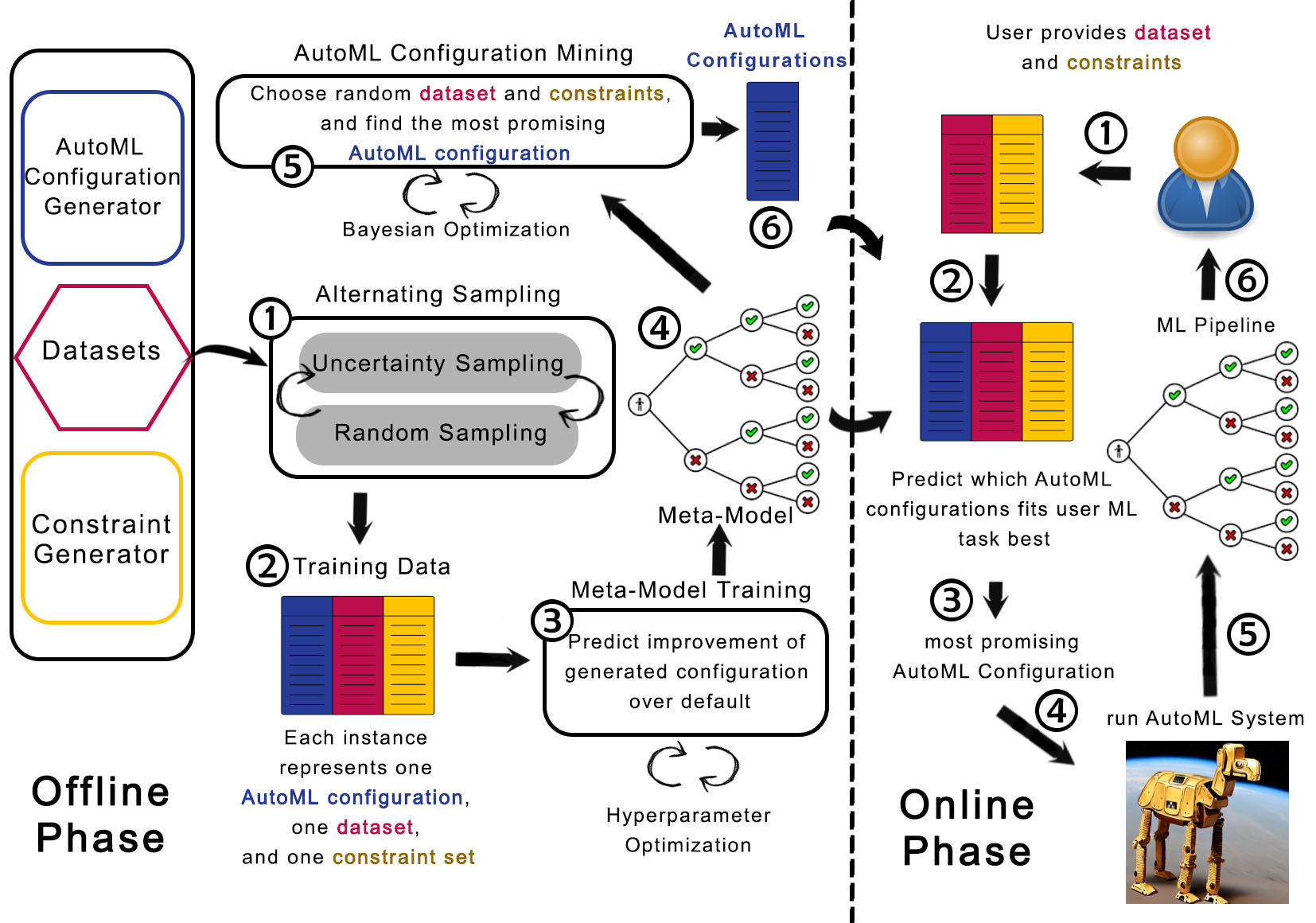}
	\caption{System Architecture of \system{}.}
	\label{figure:workflow}
\end{figure*}

\subsection{Constrained-Driven AutoML Problem}
The problem that we address in this paper is to find the parameters $\omega$ of a given AutoML system to efficiently find an ML pipeline that adheres to all user-specified constraints and achieves the highest predictive performance for a specified ML task.
Formally, 
\begin{equation}
\max\limits_{\omega \in \Omega} m(\omega) \text{ s.t. } \forall c_{i} \leq t_{i} , i \in [0,n] \label{eq:obj}
\end{equation}
where $\omega$ is a vector representing an AutoML system's own configuration; $m(\omega)$ is the average validation loss of the final ML model $\hat{f}$ returned by the AutoML system; $c_{i}$ are the constraints, and $t_{i}$ are the user-specified constraint thresholds, i.e., $\text{search time} \leq 5min$ or ML pipeline size~$\leq 1$~MB.
\newvv{For constraints, we distinguish between search constraints and ML application constraints. Search constraints concern the AutoML search process, such as search time, search main memory, and evaluation time, and ML application constraints concern the final ML pipeline, such as training/inference time, and fairness.}

Although optimizers with implicit learning of these unknown constraints can be used, we hypothesize that zero-shot adjusting of the AutoML system's own parameters (including the configuration space $A \times \Lambda$) will address this problem efficiently. 

Choosing the AutoML configuration based on a specified dataset and constraints is challenging because both the solution space (possible AutoML configurations) as well as the task space (possible datasets and constraint thresholds) are huge. Any change in any of these components might affect the final predictive performance. The nondeterminism of both ML and AutoML further aggravates these challenges.

Figure~\ref{figure:parametertypes} illustrates how constraint-driven AutoML impacts the configurations. Instead of using the default AutoML configuration, our system automatically adapts its AutoML parameters to the user-specified \emph{ML task}, which is defined by the dataset and constraints at hand. In this example, several classification methods are excluded as they are expected not to meet the specific constraint (marked red in the Dynamic AutoML Configuration).
Then, the dynamically configured AutoML system searches for ML pipelines based on the remaining ML hyperparameter search space. Finally, the \emph{model parameters} are fit to the dataset, e.g., SVM tunes the weights~$w$. Previously disabled hyperparameters either remain disabled if irrelevant or are set to default if required. For example, the dynamic AutoML configuration excluded the regularization parameter of the SVM model. However, as the final pipeline uses SVM, it will simply use the default parameter here.

%% file: 03_system.tex
\section{Constraint-Driven AutoML}
\label{sec:camlmain}

To meta-learn AutoML's own parameters $\omega$, we propose \system{}, illustrated in Figure~\ref{figure:workflow}. 
Given a user-specified dataset, search constraints, and ML-application constraints, \system{} decides which AutoML configuration - namely, which ML hyperparameter space, search strategy, and validation strategy - a given AutoML system should search to yield an ML pipeline with high predictive performance. The workflow consists of an offline and an online phase.


\newvv{The \textbf{offline phase} consists of three main steps: training data generation, meta-model training, and AutoML configuration mining. 
As input, the AutoML system engineer has to provide the AutoML space and the constraint space via generators. 
\answerVRtwo{In this paper, we benchmark training/inference time, pipeline size, and equal opportunity constraints. The engineer can extend this constraint set depending on the ML application's needs. 
Further, \system{} requires a repository of datasets. Meta-learning performs better if the user-provided datasets are similar to the datasets that are present in the repository. So there are two possible approaches to create the repository. Inside organizations, one could resort to the own history of datasets that were used in prior data science pipelines. Other than that one should create the repository with datasets that differ in dimensions, such as the number of instances, features, classes, missing values, and feature types. There are already public repositories that to some degree fulfill this diverse requirements. Following prior studies, we collected the datasets for our benchmark repository from platforms, such as OpenML~\cite{OpenML2013}, UCI ML Repository~\cite{UCI}, Kaggle~\cite{Kaggle}, and HuggingFace~\cite{HuggingFace}. }
\system{} leverages an alternating strategy~\ding{182} of random and uncertainty sampling to both explore and exploit the huge space of AutoML configurations, datasets, and constraints. 
Based on the resulting training data~\ding{183}, \system{} learns and optimizes the meta-model using cross-validation while ensuring cross-dataset generalization~\ding{184}.

Ideally, the meta-training would consider the best AutoML configuration for each ML task. However, identifying the best configuration for an ML task is nearly impossible as it would require testing the huge space of configurations per ML task.
As this goal is computationally infeasible, we relax our original problem formulation from Section~\ref{sec:problem} as follows:

\begin{equation}
\max\limits_{\omega \in \Omega} P(m(\omega) > m(\omega_{default})) \text{ s.t. } \forall c_{i} \leq t_{i} , i \in [0,n] \label{eq:obj_relaxed}
\end{equation}

To ensure a robust meta-learning approach, our intuition is to identify the AutoML configuration that is most likely more effective than the default AutoML configuration.
Thus, we train a meta-learning model that predicts whether a configuration that is different from the default AutoML configuration will result in better performance for a given task.

In the final step of the offline phase, \system{} leverages the meta-model~\ding{185} to search for the estimated optimal AutoML configuration for a random dataset and random constraints~\ding{186}. \system{} leverages BO to address this search problem. The result of this step is a large pool of promising AutoML configurations~\ding{187} for a diverse set of use cases.

In the \textbf{online phase}, the user specifies the dataset and the constraints~\ding{182}. To prepare them for the meta-model training, we encode both the dataset and the constraints in the meta-feature representation~(see Section~\ref{subsub:rep}) and combine them with the mined AutoML configurations~\ding{183}. Then, \system{} leverages the meta-model to predict which of the mined AutoML configurations fits the user-specified dataset and constraints best~\ding{184}. Then, \system{} equips the AutoML system with the resulting AutoML configuration~\ding{185} and executes it~\ding{186} with the specified search constraints. Finally, the AutoML system returns an ML pipeline that satisfies all ML application constraints~\ding{187}.
}


\subsection{Training Data for Meta-learning}
We propose active meta-learning - an approach to efficiently apply meta-learning in a scenario where the corresponding training data, both instances and labels, do not exist and need to be generated; A meta-training instance comprises a combination of a dataset, an AutoML configuration, and constraints. The label of such a training instance should specify how fitting or successful generated AutoML parameters are. 
The meta-model should learn from a set of such training instances whether a generated configuration leads to better performance than the default AutoML configuration.

To train such a meta-model, we have to answer the following questions: How do we generate the labels? How can we effectively gather training data? 
How do we encode an AutoML run as meta-features? 

\subsubsection{Meta-Target Label} \label{sec:labeling_objective}
To learn which AutoML configurations are promising, we need a meta-training dataset with prediction labels for previous AutoML runs. We need to define what \emph{success} means for a given AutoML run. 
We cannot simply choose the predictive performance as a label for an AutoML run, because the performance lives on different scales depending on the ML task at hand. Some ML tasks are harder to solve because some constraints are very restrictive. For instance, the constraint ``ML pipeline size $\leq$ 5KB'' is more restrictive than ``ML pipeline size $\leq$ 500MB'', leading to different optimally achievable prediction performance values.
Therefore, we have to find a metric that considers the entire context of an ML task as an anchor point. 
To provide such an anchor point, we run the AutoML system with default configuration as a baseline during meta-learning. The default AutoML configuration uses the full ML hyperparameter search space and the default AutoML parameters, such as hold-out validation with 33\% validation data. 
Now, our learning task is to predict whether a generated AutoML configuration yields higher predictive performance than the default AutoML configuration for the same task. This proxy metric is independent of the performance scales and the constraint hardness. 
To account for the nondeterministic behavior of AutoML, we run the AutoML system several times (ten times in our experiments) for both the generated configuration and the default configuration. Then, we obtain the fraction of cases where the default AutoML configuration was outperformed. 
We note that this might not lead to the optimum as defined in Eq.~\ref{eq:obj}, but ensures a robust choice of an AutoML configuration, avoiding performance degradation caused by non-determinism.
To avoid unnecessary computation for unsatisfiable settings in the meta-training, we first evaluate the given AutoML configuration. If all ten runs yield no ML pipeline that satisfies the specified constraints, we do not need to evaluate the default AutoML configuration anymore. 

The meta-model for active learning is a random forest regression model that predicts the fraction of runs that the given AutoML configuration outperformed the default configuration. As shown before~\cite{ThorntonHHL13}, random forest is a well-suited model for handling large complex and structured hyperparameter spaces, see Subsection~\ref{subsub:rep}.

\begin{algorithm}
\small
\caption{Training data generation}\label{alg:uncertainty_sampling}
\begin{algorithmic}[1]
\Require $\text{AutoML system $A$}, \text{Datasets~$D$}, \text{Constraint Space~$C$}, $
$\text{AutoML parameter space~$\Omega$}, \text{Random iterations~$K$},$
$\text{Sampling time~$t$}.$
\Ensure $X, Y, \text{groups}.$

\State $X \gets \emptyset$
\State $Y \gets \emptyset$
\State $\text{groups} \gets \emptyset$

\For{$i=0 \; to \; K$} \Comment{cold start}  \label{line:random}
    \State $d, c, \omega \gets \text{random\_sample}(D, C, \Omega)$ \label{line:random_sample}
    \State $X \gets X \cup \{\text{encode}(d, c, \omega)\}$ \label{line:added}
    \State $Y \gets Y \cup \{A(d, c, \omega)\}$ \Comment{Running \system{}} \label{line:run_auto}
\EndFor \label{line:random_end}

\While{$t$ not elapsed} \Comment{alternating sampling} \label{line:al_start}
    \If{ $rand() \geq 0.5$}
        \State $\text{meta\_model.fit}(X, Y)$
        \State $d, c, \omega \gets \argmax\limits_{d \in D, c \in C, \omega \in \Omega} \sigma (\text{meta\_model.predict}($
        $\text{\hspace{17em}encode}(d, c, \omega)))$ \label{line:BO}
    \Else
        \State $d, c, \omega \gets \text{random\_sample}(D, C, \Omega)$
    \EndIf
    \State $X \gets X \cup \{\text{encode}(d, c, \omega)\}$ 
    \State $Y \gets Y \cup \{ A(d, c, \omega)\}$ \Comment{Running \system{}}
    \State $\text{groups} \gets \text{groups} \cup d$
\EndWhile
\State \Return $X, Y, \text{groups}.$
\end{algorithmic}
\end{algorithm}

\subsubsection{Alternating Sampling}


To efficiently explore the space of AutoML configurations, datasets, and constraints, we leverage active learning, specifically uncertainty sampling~\cite{settles2009active}. \newv{Similar to the approach presented by Yu et al.~\cite{DBLP:conf/aaai/YuQH16}} that reduces labeling effort for standard ML classification tasks, our system chooses and generates those meta-training instances that the meta-model is most uncertain about. \newvv{However, if we only sample ML tasks around the decision boundary of whether a given AutoML configuration outperforms the default configuration, we might miss configurations that outperform the default configuration by large margins. While we \emph{exploit} the space with uncertainty sampling, we additionally \emph{explore} it with random sampling in an alternating fashion.}

\begin{sloppypar}
Algorithm~\ref{alg:uncertainty_sampling} describes the training data generation process.
Sampling requires a repository of datasets, an AutoML system, a constraint space, and a space of AutoML parameters.
To start active learning, we need initial training instances that yield the first meta-model. \system{} chooses these first instances randomly (Lines~\ref{line:random}-\ref{line:random_end}).
In particular, \system{} randomly chooses the dataset~$d$, the constraints~$c$, and the AutoML configuration $\omega$ (Line~\ref{line:random_sample}). Then, those components are encoded as meta-features and added to the meta-training set~(Line~\ref{line:added}). The corresponding AutoML run is executed and compared with the default configuration to obtain the corresponding label~(Line~\ref{line:run_auto}).
Then, the alternating sampling process starts (Line~\ref{line:al_start}). 
The system chooses uniformly at random whether to apply random or uncertainty sampling.
Uncertainty sampling picks the most uncertain instance among all given instances. 
To find uncertain instances in this huge search space (combinations of datasets, AutoML configurations, and constraints), we leverage BO, which learns a surrogate model to predict which AutoML parameters yield high predictive performance and samples only promising instances by trading off exploration and exploitation. In Line~\ref{line:BO}, BO identifies the combination of ($d$,$c$,$\omega$) that leads to the highest standard deviation across all trees of the random forest meta-model.
We repeat this two-step loop until the time limit has been reached.
\end{sloppypar}

\subsubsection{Parallelization and Optmizations}
To speed up the presented sequential algorithm, we parallelize it asynchronously. Each worker always accesses the latest training instances. Once a new meta-training instance and a corresponding label are available, the meta-training data is locked briefly to add the new instance.
We found that the more common approach~\cite{zhang2020efficient} to predict the label for a newly sampled instance with the current meta-model and adding both to the meta-training data does not work well for our scenario. 
Our label is only predicted and is thus only an approximation of the ground truth. If the label is not correct, the search could fall into the wrong direction. 
Therefore, our approach only adds a new instance once the label is confirmed. 
To avoid the same instances being evaluated in parallel, we start each nondeterministic BO run with different seeds. As the search space is huge, it is highly unlikely that similar instances will be sampled during the same period.

\subsubsection{Meta-Feature Representation} \label{subsub:rep}
To estimate whether an AutoML configuration yields higher predictive performance than the default AutoML configuration, the meta-model has to know the dataset, the AutoML parameters, and the constraints.
We encode each of these components in a meta-feature vector.

\paragraph{Dataset Features}
\begin{sloppypar}
For encoding datasets into meta-feature vectors, multiple approaches have been proposed~\cite{MetaLearning,MetaData,feurer2015efficient}. We leverage the $32$~meta-features proposed by Feurer et al.~\cite{feurer2015efficient}, such as the class entropy, the number of features, classes, and instances.  
\end{sloppypar}

\paragraph{Constraint Features}
All constraints, such as inference time~$\leq 0.001s$, can be represented by the corresponding threshold. If the user does not specify the constraint, we set the maximum possible default value. Extending the set of constraints is always possible. The safest strategy is to train the meta-model from scratch. However, one can also leverage the assumption that the missing constraint was simply set to default. Thus, all previous training instances can be appended with the default value for the new constraint and new instances with novel thresholds for the constraint can be generated for new instances. This way, we can continue meta-training asynchronously without the need of starting from scratch.
\newvv{The same reasoning applies to extending the search space of the AutoML parameters. However, this only works, if one does not change the underlying AutoML system that we compare to - e.g. if one uses the state-of-the-art AutoML system as a comparison, one can leverage the assumption that the missing component was simply not chosen. This way, we can continue meta-training without the need of starting from scratch.}

\paragraph{AutoML Configuration Features}
\label{sec:automl_features}
To encode an AutoML configuration, we distinguish numeric parameters and categorical ones. 
Numeric AutoML parameters, such as the choice of the validation fraction, are simply added to the meta-feature vector.
We encode the ML hyperparameters as binary values. The AutoML system either optimizes each ML~hyperparameter~(\emph{True}) or uses its corresponding default value~(\emph{False}). 
For instance, the AutoML system can optimize the number of neighbors for $K$~nearest neighbors or use its default $K=5$.

We follow the well-known assumption that the ML hyperparameter space has a tree structure where each node represents an ML~hyperparameter~\cite{ThorntonHHL13,bergstra2011algorithms} and each edge represents the dependency on its parent ML hyperparameter. Figure~\ref{figure:treespace} shows a branch of this tree. We describe the details of how we structure this tree in Section~\ref{sec:adjustableAutoML}.
If we do not optimize an ML~hyperparameter higher up in the tree, we will not optimize any of its descendant ML~hyperparameters either. For instance, if we remove the $K$-nearest-neighbor classifier from the choice of possible classifiers, we also do not need to optimize the number of neighbors $k$.
We refer the reader to our repository~\cite{automl_space} for the complete tree space that we leverage.

The aforementioned set of meta-features assumes uniform hardware specifications at training and deployment time which cannot always be guaranteed. If the hardware of meta-learning training is different from the hardware where \system{} is deployed, one can apply calibration strategies that were proposed for database query optimization cost models~\cite{DBLP:conf/sigmod/GhodsniaBN14}. For instance, one could run a lightweight benchmark to understand the hardware performance difference and obtain corresponding scaling functions. 


\begin{figure}
	\centering
	\input{bench}
	\caption{\answerVRone{Mapping search time from one environment to another environment.}}
	\label{figure:bench}
\end{figure}
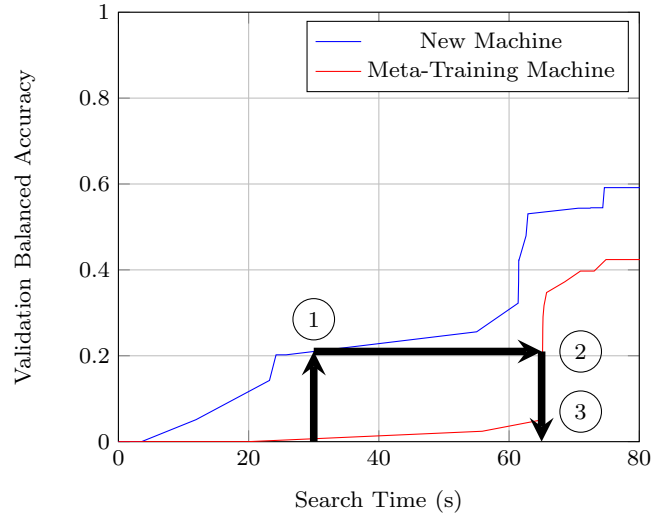

\answerVRone{We propose a simple calibration method to implement this idea. In particular, one can execute an AutoML system and keep track of the highest predictive performance on the validation set for one or multiple datasets that generally benefit from longer search times and use a performance mapping to calibrate the search time. One could implement the same idea using the test set but this would require additional computation and usually the test data is not accessible to the AutoML system during runtime.

During an offline step, the static \system{} is applied on the selected datasets on both machines, the source and the target environment, and records the improvement of validation accuracy over time. These benchmarks lead to two graphs, as shown in Figure~\ref{figure:bench}. 
During the online process for a new dataset, one can now specify a desired search time on the target machine, which will be internally mapped to a search time that achieves the same validation accuracy on the source machine. 
In Figure~\ref{figure:bench}, we marked 30 seconds on the target machine and the graph visualizes how it maps to a different search time based on the equality of the validation accuracy.
\system{} searches for the search time where the meta-training machine reached this validation accuracy and uses the adjusted search time to configure the AutoML parameters for the new machine. Note that it still runs only 30 seconds on the target machine but sets the configuration space based on the adjusted search time.
The advantage of this calibration method is that it works for any hardware setup without the requirement of obtaining hardware meta-information. To improve the reliability of the calibration, one should conduct multiple runs and average the results.
The approach will be costly if the targeted search times are rather high. However, we argue that in these cases calibration is not necessary as the search time is long enough. This is also validated by our experiments discussed in Section~\ref{sec:hardwareajustmentexperiment}.
}

\subsection{Meta-Model Training}
\begin{sloppypar}
Once the meta-data sampling is finished, \system{} trains the final meta-model.
The straightforward approach would be to use the same model that was trained for uncertainty sampling. However, this model is suboptimal because it might be overfitted to certain datasets that are more frequently sampled than others due to their uncertainty estimation. 
Further, we do not optimize the model hyperparameters during uncertainty sampling as it would significantly slow down the training data generation.
For these reasons, we apply hyperparameter optimization on the meta-model after sampling has finished with 10-fold time series cross-validation because active learning sampling creates the dataset incrementally.

To achieve optimal performance, we train two meta-models, one for AutoML configuration mining and one to rank the large pool of mined AutoML configurations.

For AutoML configuration mining, we use the same objective as for the surrogate model for uncertainty sampling~(see Section~\ref{sec:labeling_objective}): we predict the fraction of runs that the given AutoML configuration outperformed the default one (regression). 
For ranking the mined AutoML configurations, we predict whether the given AutoML configuration outperforms the default one at least once (classification).
The regression meta-model contains more information than the classification meta-model because it estimates how much better the given AutoML configuration is compared to the default one whereas the classification model estimates only whether the AutoML configuration is better than the default one. However, as the regression task is much harder than the classification task, the regression meta-model is more likely to make mistakes and therefore more unstable. Yet, as we describe in Section~\ref{sec:mining}, we query the regression meta-model many times, avoid local optima/mistakes, and converge over time to a well-performing AutoML configuration.
In turn, we only query the ranking meta-model once. Therefore, we need to make sure that it makes no mistakes and is as conservative as possible. This way, we ensure that the highest ranked AutoML configuration is robust - meaning it outperforms at least the default configuration. 
\end{sloppypar}

\subsection{AutoML Configuration Mining} \label{sec:mining}
Given an ML task and a generated configuration, the trained regression meta-model can predict whether the generated configuration will be more effective than the default configuration or not. The question is how we can leverage this regression meta-model to find the best AutoML configuration for a new dataset and user-specified constraints.
To use the trained regression meta-model, we need a set of generated candidate configurations for each of which we can carry out the inference. Here, we are looking for the AutoML configuration that yields the best predictive performance for a given dataset and given constraints. 

\newvv{The simplest approach would be to generate a large number of random configuration candidates and let the regression meta-model predict which of these configurations has the highest likelihood of success. The disadvantage of this approach is that many of the randomly generated configurations will perform poorly and we cannot generate all possible configurations. The advantage of this approach is that the generation of these random configurations can be performed in the offline phase. During the online phase, we would only apply inference. The cost of inference is minimal - e.g. predicting one million configurations takes around 1s.

Instead of random sampling, we could also apply BO. We could maximize the estimated likelihood that a generated configuration outperforms the default configuration, and freeze all meta-features for the user-specified dataset and constraints: 

\begin{equation}
\hat{\omega} \gets \argmax\limits_{\omega \in \Omega} \text{meta\_model.predict}(\text{encode}(d, c, \omega)) \label{eq:probing}
\end{equation}

The advantage of BO is that it would adjust the configuration to the specified dataset and constraints. The disadvantage of BO is that it is slow. For instance, performing 1000 iterations would take more than 700s. Waiting for more than 10min before we even start the AutoML system is not viable - especially if the user is interested in fast development cycles.

We propose a hybrid approach that combines the strengths of both probing strategies.
In the offline phase, we randomly sample a dataset and constraints - similar to Line~\ref{line:random_sample}. But instead of randomly sampling a configuration~$\omega$, we leverage BO to find the most promising configuration for this randomly generated ML task with the help of the regression meta-model. This way, we generate a large number of promising random configurations offline. In the online phase, we let the classification meta-model choose which of these promising random configurations fits the specified dataset and constraints best. 
Then, \system{} sets up the actual AutoML system with this configuration and executes it.
}

\begin{figure}
	\centering
	{\scriptsize
    \begin{forest} for tree={
        grow=east
    }
    [
    	[search
    		[time-related
    		    [Search Time]
    		    [Evaluation Time]
    		]
    		[hardware-related
    		    [Memory]
    		    [Parallelism]
    		]
    		[AutoML system-specific
    		    [Ensemble Size]
    		    [Search Space]
    		]
    	]
    	[ML application
    	    [known
    	        [Privacy]
    	        [Interpretability
    	            [$|\text{Features}|$]
    	            [ML Pipeline]
    	        ]
    	    ]
    		[unknown
    		    [Efficiency
        		    [Training Time]
        		    [Inference Time]
        		    [ML Pipeline Size]
        		]
    		    [Fairness]
    		    [Robustness]
    		    [Correctness]
    		    [Security]
    		]
    	]
    ]
    \end{forest}
}
	\caption{AutoML constraints.}
	\label{figure:constraints_taxonomy}
	\vspace{-1.0em}
\end{figure}
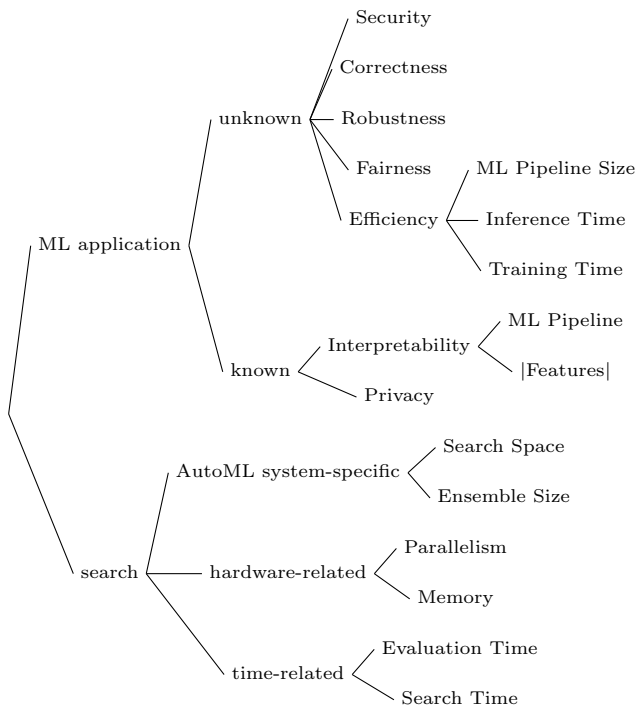

\subsection{AutoML Parameters}
\label{sec:adjustableAutoML}

\begin{sloppypar}
Adapting AutoML parameters is only meaningful if there is a wide range of parameters that are in fact adaptable. 
In contrast to Auto-Sklearn and AutoGluon, we implemented \system{} to not only provide access to the common user-adjustable AutoML parameters, such as whether to use ensembling, incremental training, or which validation strategy, but also to allow external adjustment of every single ML hyperparameter in the search space. This way, it can be dynamically decided whether those parameters should be optimized or not, as shown in Figure~\ref{figure:parametertypes}.

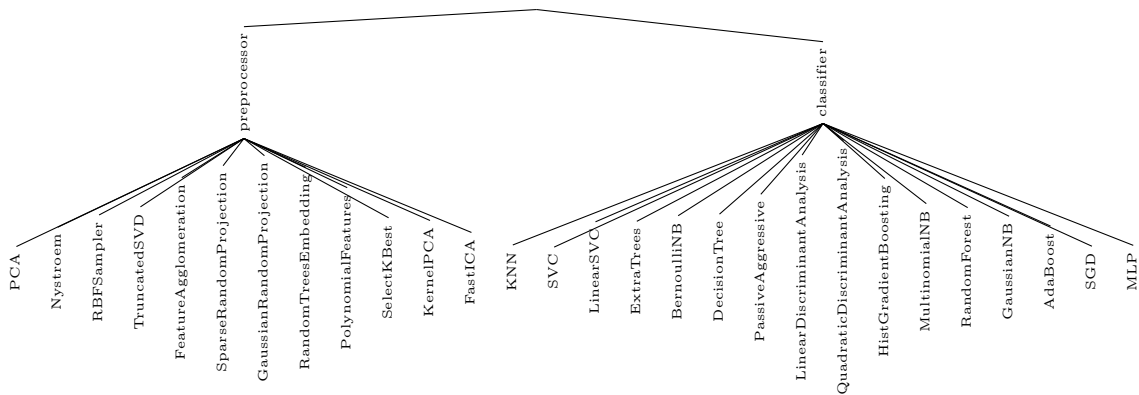
\begin{figure*}
	\centering
	{\tiny
    \begin{forest} for tree={
        grow=south
    }
    [
    	[preprocessor,rotate=90
    		[PCA,rotate=90]
    		[Nystroem,rotate=90]
    		[RBFSampler,rotate=90]
    		[TruncatedSVD,rotate=90]
    		[FeatureAgglomeration,rotate=90]
    		[SparseRandomProjection,rotate=90]
    		[GaussianRandomProjection,rotate=90]
    		[RandomTreesEmbedding,rotate=90]
    		[PolynomialFeatures,rotate=90]
    		[SelectKBest,rotate=90]
    		[KernelPCA,rotate=90]
    		[FastICA,rotate=90]
    	]
    	[classifier,rotate=90
    		[KNN,rotate=90]
            [SVC,rotate=90]
            [LinearSVC,rotate=90]
            [ExtraTrees,rotate=90]
            [BernoulliNB,rotate=90]
            [DecisionTree,rotate=90]
            [PassiveAggressive,rotate=90]
            [LinearDiscriminantAnalysis,rotate=90]
            [QuadraticDiscriminantAnalysis,rotate=90]
            [HistGradientBoosting,rotate=90]
            [MultinomialNB,rotate=90]
            [RandomForest,rotate=90]
            [GaussianNB,rotate=90]
            [AdaBoost,rotate=90]
            [SGD,rotate=90]
            [MLP,rotate=90]
    	]
    ]
    \end{forest}
}
	\caption{Slice of the tree space that we use in our implementation.}
	\label{figure:treespace}
	\vspace{-1.0em}
\end{figure*}

We extend the ML hyperparameter space of Auto-Sklearn~\cite{feurer2015efficient} additionally supporting oversampling strategies random oversampling, SMOTE~\cite{SMOTE}, and ADASYN~\cite{DBLP:conf/ijcnn/HeBGL08} to address class imbalance. Further, we added support for one-vs-rest classification to improve multi-class classification. 
We refer the reader to our repository~\cite{automl_space} for the complete tree space that we leverage. 
We structure the ML hyperparameter space in a tree~\cite{automl_space}, as proposed in Auto-Weka~\cite{ThorntonHHL13}. Figure~\ref{figure:treespace} represents a slice of the leveraged tree space. 
The first level of the tree contains all main components of the ML pipeline: categorical encoding, imputation, scaling, classifier, feature preprocessor, augmentation, sampling, and class weighting.
Below this level, each component can be implemented by various strategies and each strategy has its own hyperparameters. 
This way, the ML hyperparameter space naturally builds up a tree.
The hierarchical organization of the ML hyperparameter space is essential to allow the meta-model to prune a large part of the ML hyperparameter space as early as possible. This way, the AutoML system will not optimize the child ML~hyperparameters if their parent ML~hyperparameter is not optimized. Instead, the system will use their default value. For instance, by providing a hierarchical structure, we allow the meta-model to realize that no preprocessing transformation will be beneficial for a specific setting, instead of deciding for every single preprocessor and all its corresponding hyperparameters whether to optimize it or not.
\end{sloppypar}

\subsection{Constraints}
\label{sec:constraintsdecription}
\begin{sloppypar}
In constraint-driven AutoML, the user can define constraints, which might concern the AutoML process or the ML application, as shown in Figure~\ref{figure:constraints_taxonomy}.

\noindent\textbf{Search constraints} limit time-related, hardware-related, or system-specific aspects of the AutoML process.
Time-related search constraints limit the search time or the evaluation time. 
Hardware-related search constraints are limits on the memory or parallelism.
System-specific search constraints are limits on the size of ensembles or the search space.

The most important search constraint limits the \emph{search time}. This search constraint is mandatory for each AutoML run and therefore it represents the class of search constraints well. 
For fast development cycles, data scientists will limit the search time to less than an hour to quickly experiment with the pipeline.

\noindent\textbf{ML application constraints} restrict the ML pipelines with regard to different quality dimensions. 
Zhang et al.~\cite{zhang20testing} described  7 quality dimensions that can serve as constraints: correctness, robustness, security, privacy, efficiency, fairness, and interpretability. These constraints can be categorized along two dimensions. 

Gelbart et al.~\cite{spearmint} differentiate between unknown and known constraints as also illustrated in our constraint taxonomy. \emph{Known constraints} are those constraints that can be checked before training and evaluating a model. For instance, knowing that an $\varepsilon$-differentially private implementation of classifiers~\cite{chaudhuri2011differentially} is used apriori ensures that privacy constraints are satisfied.
Another example of known constraints is a restriction with respect to the ML pipeline components or the number of features to improve the interpretability of the resulting ML pipeline.
In contrast, \emph{unknown constraints} refer to those that can only be checked once the model is trained and evaluated. For instance, most efficiency constraints have this property.

Generally, our approach can integrate any known constraint easily by adding an if~statement at the beginning of the objective function. 
For our experiments, we focus on unknown constraints.

The second dimension along which one can differentiate constraints refers to their dependence on the ML pipeline and/or the data. For our experiments, we focus on constraints that significantly depend on the pipeline and not so much on the dataset. To incorporate more dataset-dependent constraints, such as fairness one would need to use more dataset-specific meta-features in the meta-model.

All in all, among the seven quality dimensions proposed by Zhang et al.~\cite{zhang20testing}, we focus on correctness, efficiency, and fairness. In particular, we always maximize correctness i.e. the predictive performance. Further, we choose three well-established efficiency constraints \emph{training time}, \emph{inference time}, and \emph{ML pipeline size}\footnote{For some ML models, such as random forest and KNN, the model size is data dependent.}\newvv{, and equal opportunity~\cite{hardt2016equality} which is a fairness measure}. All four are \emph{unknown} constraints and depend on the ML pipeline.

The relevance of the three efficiency constraints is particularly high in edge computing and streaming scenarios.
In streaming scenarios, reducing inference time is vital to ensure continuous real-time predictions. As the data is evolving, the model requires constant retraining. In continuous training scenarios, enforcing training time limits plays a significant role. The same constraint type is relevant for federated learning~\cite{FederatedLearning}, where users continue training on their own devices.
Finally, to apply ML on IoT devices or smartphones, it is important to limit memory consumption.
\end{sloppypar}
\answerVRtwo{
\subsection{Extending the list of constraints}
\label{sec:extendconstraints}
First, one has to define the user-defined function that describes the constraint. The process depends on whether we want to create an ML application or search constraint. 

For ML application constraints, one has to implement the following template:}

\begin{lstlisting}[language=Python]
def constraint(pipeline, training_time, X_train, y_train, X_val, y_val, threshold, constraint_specific): True/False
\end{lstlisting}

\answerVRtwo{This function takes the trained pipeline and its training time, the split data, the constraint threshold, and constraint-specific parameters, such as the sensitive attribute for fairness. The output of this function is whether the given ML pipeline passes the constraint or not.

After implementing the user-defined function, one has to add a new feature to the metadata representation and continue meta-training. To account for the new metadata feature, first one has to retrain the meta-model. With the retrained meta-model, one can continue alternating sampling including the new constraint.
Finally, one has to generate additional configurations that also cover the new constraint as described in Section~\ref{sec:mining}.

For search constraints, one has to additionally implement an initialize function that starts the measuring at the beginning of search and another function that checks whether the search constraint is still satisfied.
}

\subsection{Constrained Optimization} \label{sec:constrainedhpo}
So far we know how to train the meta-learning approach and how to retrieve an adapted AutoML configuration dynamically. Now, we explain how \system{} optimizes the ML hyperparameters under constraints. Previous systems by default consider the predictive performance as the objective function, which is not sufficient and requires adjustment. Furthermore, aspects such as ensembling have to be adjusted as we need to make sure that only constraint-satisfying models are ensembled and that the final ensemble also satisfies the constraints. 

To support ML application constraints we formulate the objective function as follows for \system{}:

\begin{equation*}
    \max \left( -1 \cdot \left(\sum_{i=1}^{n} \Delta{}c_{i}\right) + \left(\left[\sum_{i=1}^{n} \Delta{}c_{i} == 0 \right] \cdot BA \right)\right), 
\end{equation*}
\normalsize

\begin{sloppypar}
where $\Delta{}c_{i}$ is the distance to satisfying the $i$th constraint and $BA$ is balanced accuracy. This objective ensures to satisfy the constraints first and then optimizes the balanced accuracy. This way, the user can set thresholds for any of the supported constraints through an API.
As the BO framework to maximize this objective, we choose Optuna~\cite{optuna_2019}, which leverages the tree-structured Parzen estimator~(TPE) as the surrogate model. TPE is well-suited for our tree-structured ML search space.

To enable model ensembling in \system{}, we integrate the greedy ensembling strategy proposed by Caruana et al.~\cite{DBLP:conf/icml/CaruanaNCK04}. The strategy iteratively adds the model that maximizes ensemble validation predictive performance as long as all constraints are satisfied.

To enable hyperparameter optimization on large data, we implement incremental training similar to successive halving~\cite{DBLP:journals/jmlr/LiJDRT17}. First, we train a model on a small sample containing 10 instances per class. Then, we double the training set size and train the model again. We continue this approach until either the evaluation time is over or the ML hyperparameter configuration is pruned because it performed worse than the median configuration of the history. Further, for constraint metrics that monotonically increase with the training set size, such as the training time or ML pipeline size, we stop the configuration evaluation as early as possible to avoid unnecessary computation. As incremental training might result in a large number of ML hyperparameter evaluations, the danger of overfitting increases. L{\'e}vesque proposes to reshuffle the validation split after each evaluation to avoid overfitting~\cite{levesque2018bayesian}. Therefore, we implemented this option in \system{} as well and expose it as an AutoML parameter.
\end{sloppypar}

%% file: bench.tex
\begin{tikzpicture}[scale=1.0]

\begin{axis}[
  ymin=0, ymax=1,
        xmin=0,xmax=80,
        grid=both,
        xlabel=Search Time (s),
        ylabel=Validation Balanced Accuracy,
        legend entries={
	New Machine, Meta-Training Machine}]

\addplot[color=blue] coordinates {(0.0,0.0)(2.033047676086426,0.0)(2.3239872455596924,0.0)(3.5498836040496826,0.0)(11.92128038406372,0.05121212121212121)(23.202535152435303,0.14257575757575758)(24.21026039123535,0.20218855218855217)(25.75804114341736,0.20218855218855217)(55.02366232872009,0.25582491582491584)(61.39332866668701,0.3225084175084175)(61.464417695999146,0.3725084175084175)(61.48668718338013,0.4226430976430976)(61.519335985183716,0.4226430976430976)(62.6049439907074,0.47932659932659927)(62.8947274684906,0.5305892255892255)(70.62122464179993,0.5436195286195284)(72.45507478713989,0.5436700336700336)(72.55817794799805,0.5446632996632995)(74.38974905014038,0.5446632996632995)(74.64057779312134,0.5915488215488215)(76.55340218544006,0.5915488215488215)(83.82175135612488,0.5915488215488215)(85.31370854377747,0.5915488215488215)(86.21704292297363,0.5915488215488215)(86.33582758903503,0.5915488215488215)(87.42698073387146,0.5915488215488215)(88.80031323432922,0.5915488215488215)(89.02108073234558,0.5915488215488215)(90.12813568115234,0.5915488215488215)(91.67118620872498,0.6113973063973064)(93.20859599113464,0.6779966329966329)(96.95383644104004,0.6779966329966329)(99.65717434883118,0.704107744107744)(102.59631872177124,0.704107744107744)(103.9609010219574,0.704107744107744)(105.38484311103821,0.704107744107744)(106.7212450504303,0.704107744107744)(108.45074152946472,0.704107744107744)(108.75419163703918,0.704107744107744)(111.30226540565491,0.704107744107744)(115.64438557624817,0.7198484848484848)(115.730055809021,0.7223232323232323)(123.33126068115234,0.7223232323232323)(124.33214235305786,0.7223232323232323)(124.84549498558044,0.7223232323232323)(127.25406694412231,0.7223232323232323)(127.78285312652588,0.7332659932659932)(131.18618273735046,0.7372390572390571)(132.91199326515198,0.7372390572390571)(133.05547547340393,0.7547643097643097)(134.62130784988403,0.7547643097643097)(137.55916905403137,0.7547643097643097)(137.95606112480164,0.7547643097643097)(141.4463620185852,0.7547643097643097)(141.45980048179626,0.7547643097643097)(145.7315113544464,0.7547643097643097)(146.64138674736023,0.7547643097643097)(147.58426475524902,0.7547643097643097)(148.0317840576172,0.7547643097643097)(149.19660782814026,0.7547643097643097)(149.36277270317078,0.7547643097643097)(149.68565940856934,0.7547643097643097)(149.85610127449036,0.7831986531986532)(150.2094690799713,0.7831986531986532)(150.92695689201355,0.7831986531986532)(151.312420129776,0.7831986531986532)(151.3232865333557,0.7831986531986532)(152.72881436347961,0.7831986531986532)(152.87612175941467,0.7831986531986532)(153.18546175956726,0.7831986531986532)(154.23961305618286,0.7831986531986532)(154.3803150653839,0.7831986531986532)(157.47031044960022,0.7831986531986532)(157.67682123184204,0.7831986531986532)(158.16354036331177,0.7831986531986532)(158.29116201400757,0.7831986531986532)(159.3491084575653,0.7831986531986532)(159.63826870918274,0.7831986531986532)(163.9053611755371,0.7831986531986532)(164.51724195480347,0.7831986531986532)(166.04026126861572,0.7831986531986532)(169.2346339225769,0.7831986531986532)(173.8005714416504,0.7831986531986532)(176.31486105918884,0.7965488215488217)(177.84497690200806,0.7965488215488217)(180.20277667045593,0.7965488215488217)(183.02606558799744,0.7965488215488217)(184.6558928489685,0.7965488215488217)(185.47928524017334,0.7965488215488217)(186.46968603134155,0.7965488215488217)(188.63443779945374,0.7965488215488217)(202.0214250087738,0.8179124579124579)(202.38558626174927,0.8179124579124579)(204.2343020439148,0.8179124579124579)(205.54936718940735,0.8179124579124579)(210.8171741962433,0.8179124579124579)(211.61928176879883,0.8179124579124579)(212.09466457366943,0.8179124579124579)(212.49084305763245,0.8179124579124579)(213.4697117805481,0.8179124579124579)(214.1851327419281,0.8179124579124579)(214.83352375030518,0.8179124579124579)(215.44372630119324,0.8179124579124579)(215.7865207195282,0.8179124579124579)(218.34287810325623,0.827946127946128)(220.50623226165771,0.827946127946128)(222.21777081489563,0.827946127946128)(222.85044407844543,0.827946127946128)(224.1385190486908,0.827946127946128)(227.819965839386,0.827946127946128)(227.89406561851501,0.827946127946128)(229.7202398777008,0.827946127946128)(230.37264227867126,0.827946127946128)(231.7475550174713,0.827946127946128)(232.82073068618774,0.827946127946128)(232.85056471824646,0.827946127946128)(234.2287678718567,0.827946127946128)(235.55018663406372,0.827946127946128)(239.32838606834412,0.827946127946128)(243.04386234283447,0.827946127946128)(249.2045247554779,0.838080808080808)(254.3872196674347,0.838080808080808)(256.1482753753662,0.838080808080808)(264.8446297645569,0.838080808080808)(266.2132251262665,0.855841750841751)(266.2707064151764,0.855841750841751)(267.61489391326904,0.855841750841751)(271.3288083076477,0.855841750841751)(272.19345259666443,0.855841750841751)(274.0496246814728,0.8622895622895623)(276.1881754398346,0.8622895622895623)(278.64550971984863,0.8622895622895623)(281.2573471069336,0.8622895622895623)(281.4927475452423,0.8622895622895623)(283.4417932033539,0.888063973063973)(288.43639516830444,0.888063973063973)(290.32013607025146,0.888063973063973)(291.2061975002289,0.888063973063973)(292.1715204715729,0.888063973063973)(292.2230796813965,0.888063973063973)(293.5981900691986,0.888063973063973)(294.0634934902191,0.888063973063973)(295.42092967033386,0.888063973063973)(297.33304810523987,0.888063973063973)(299.07822489738464,0.888063973063973)(307.0713052749634,0.888063973063973)(310.79979848861694,0.888063973063973)(313.0718140602112,0.888063973063973)(314.4081964492798,0.888063973063973)(327.2380578517914,0.888063973063973)(331.0197124481201,0.888063973063973)(332.8995318412781,0.888063973063973)(333.47310495376587,0.888063973063973)(334.4168689250946,0.888063973063973)(334.6857671737671,0.8886026936026937)(336.3704423904419,0.8886026936026937)(336.417564868927,0.8886026936026937)(338.88472151756287,0.8886026936026937)(339.9262161254883,0.8886026936026937)(340.74109983444214,0.8886026936026937)(341.7944233417511,0.8886026936026937)(342.59589743614197,0.8886026936026937)(344.14688086509705,0.8886026936026937)(347.5573399066925,0.8886026936026937)(347.8272924423218,0.8886026936026937)(348.49040961265564,0.8886026936026937)(348.99312710762024,0.8886026936026937)(350.30998611450195,0.8886026936026937)(351.9231901168823,0.8886026936026937)(352.7724678516388,0.8886026936026937)(352.79497170448303,0.8886026936026937)(353.33895111083984,0.8886026936026937)(354.12458968162537,0.8886026936026937)(354.1905996799469,0.8886026936026937)(356.0160491466522,0.8886026936026937)(357.50425028800964,0.8886026936026937)(359.72138595581055,0.8886026936026937)(363.9041209220886,0.8901683501683502)(369.8053021430969,0.8901683501683502)(371.1606547832489,0.8901683501683502)(372.24172616004944,0.8901683501683502)(375.0483958721161,0.8901683501683502)(377.4880430698395,0.8901683501683502)(378.9200990200043,0.8901683501683502)(380.8710768222809,0.8901683501683502)(381.7625365257263,0.8901683501683502)(382.297199010849,0.8901683501683502)(383.2581100463867,0.8901683501683502)(391.38341307640076,0.8901683501683502)(394.2019922733307,0.8901683501683502)(394.4609591960907,0.8901683501683502)(396.16406178474426,0.8901683501683502)(399.56561040878296,0.8901683501683502)(400.04369711875916,0.8901683501683502)(400.9907441139221,0.8901683501683502)(402.3568880558014,0.8901683501683502)(403.19039368629456,0.8901683501683502)(404.32932567596436,0.8901683501683502)(404.5750608444214,0.8901683501683502)(405.88888335227966,0.8901683501683502)(406.14354729652405,0.8901683501683502)(406.37784600257874,0.8901683501683502)(407.61981868743896,0.8901683501683502)(409.08398389816284,0.8901683501683502)(410.90261125564575,0.8901683501683502)(411.3098318576813,0.8901683501683502)(411.3963074684143,0.8901683501683502)(414.77733159065247,0.8916666666666666)(415.83122301101685,0.8916666666666666)(416.50339698791504,0.8916666666666666)(416.625830411911,0.8916666666666666)(416.6313304901123,0.8916666666666666)(416.9233627319336,0.8916666666666666)(418.11177372932434,0.8916666666666666)(418.1369616985321,0.8916666666666666)(419.9227035045624,0.8916666666666666)(420.3751084804535,0.8916666666666666)(421.22745633125305,0.8916666666666666)(422.99761414527893,0.8916666666666666)(424.3070492744446,0.8916666666666666)(425.61901235580444,0.8916666666666666)(425.90281200408936,0.8916666666666666)(427.80849742889404,0.8916666666666666)(428.26202917099,0.8916666666666666)(430.15433382987976,0.8916666666666666)(430.15968680381775,0.8916666666666666)(431.8187680244446,0.8916666666666666)(433.21560287475586,0.8916666666666666)(435.73509430885315,0.8916666666666666)(437.1072793006897,0.8916666666666666)(439.4502592086792,0.8916666666666666)(441.1286494731903,0.8916666666666666)(445.49162220954895,0.8916666666666666)(448.8542230129242,0.9018855218855218)(449.1763668060303,0.9018855218855218)(450.7684369087219,0.9018855218855218)(453.16319036483765,0.9018855218855218)(455.91576051712036,0.9018855218855218)(456.93940806388855,0.9018855218855218)(460.64317417144775,0.901969696969697)(463.73065543174744,0.901969696969697)(466.5814628601074,0.901969696969697)(469.66939187049866,0.901969696969697)(472.07781982421875,0.9036363636363636)(476.4055771827698,0.9036363636363636)(476.96750712394714,0.9131481481481482)(477.24979066848755,0.9131481481481482)(477.5086889266968,0.9131481481481482)(478.50889587402344,0.9131481481481482)(478.7803637981415,0.9131481481481482)(478.81771183013916,0.9131481481481482)(479.95713806152344,0.9131481481481482)(480.882319688797,0.9131481481481482)(482.5763874053955,0.9131481481481482)(485.50644969940186,0.9131481481481482)(491.50739884376526,0.9131481481481482)(496.0201003551483,0.913888888888889)(498.3639929294586,0.913888888888889)(499.281973361969,0.913888888888889)(500.88765239715576,0.9182491582491583)(503.4158754348755,0.9182491582491583)(504.83131170272827,0.9182491582491583)(506.2140872478485,0.9182491582491583)(507.5659577846527,0.9182491582491583)(509.3954029083252,0.9182491582491583)(513.7434847354889,0.9182491582491583)(517.5066621303558,0.9182491582491583)(518.8620777130127,0.9182491582491583)(519.6876170635223,0.9182491582491583)(520.143877029419,0.9182491582491583)(522.4762096405029,0.9182491582491583)(524.1832983493805,0.9182491582491583)(524.4944944381714,0.9183501683501685)(528.6535782814026,0.9183501683501685)(529.444760799408,0.9183501683501685)(532.7390396595001,0.9183501683501685)(532.8459804058075,0.9200168350168351)(533.0573575496674,0.9200168350168351)(534.4089307785034,0.9200168350168351)(536.1104454994202,0.9200168350168351)(543.2184777259827,0.9200168350168351)(544.3169226646423,0.9200168350168351)(546.51194190979,0.9200168350168351)(547.0476591587067,0.9200168350168351)(548.4979295730591,0.9200168350168351)(549.5973498821259,0.9200168350168351)(550.9456295967102,0.9200168350168351)(552.2271130084991,0.9200168350168351)(554.8393058776855,0.9200168350168351)(556.622044801712,0.9200168350168351)(559.0149631500244,0.9200168350168351)(559.5681319236755,0.9200168350168351)(560.3317973613739,0.9200168350168351)(560.8927974700928,0.9200168350168351)(563.0353286266327,0.9200168350168351)(563.6018173694611,0.9200168350168351)(564.2414467334747,0.9200168350168351)(564.7236704826355,0.9200168350168351)(565.8575830459595,0.9200168350168351)(566.3633878231049,0.9200168350168351)(568.6652820110321,0.9200168350168351)(569.9458951950073,0.9200168350168351)(570.0650672912598,0.9200168350168351)(571.7288000583649,0.9200168350168351)(573.701996088028,0.9200168350168351)(574.0702321529388,0.9200168350168351)(576.445098400116,0.9200168350168351)(578.6416692733765,0.9200168350168351)(581.2523913383484,0.9200168350168351)(582.3863427639008,0.9200168350168351)(584.268857717514,0.9200168350168351)(586.9971415996552,0.9200168350168351)(591.2614090442657,0.9200168350168351)(593.5984590053558,0.9200168350168351)(594.3815495967865,0.9200168350168351)(598.2121813297272,0.9200168350168351)(599.1797108650208,0.9200168350168351)(599.2545609474182,0.9200168350168351)(599.2673645019531,0.9200168350168351)(599.5344302654266,0.9200168350168351)(599.5907559394836,0.9200168350168351)(599.6176261901855,0.9200168350168351)(599.8594448566437,0.9200168350168351)(599.8888463973999,0.9200168350168351)(599.8902697563171,0.9200168350168351)
};

\addplot[color=red] coordinates {(0.0,0.0)(6.226100206375122,0.0)(6.828287601470947,0.0)(7.1872570514678955,0.0)(12.61104702949524,0.0)(18.86361050605774,0.0)(19.923370122909546,0.0)(55.88178586959839,0.024217171717171717)(64.51039671897888,0.049217171717171715)(64.81298589706421,0.0754124579124579)(64.8168454170227,0.10103535353535353)(65.0459246635437,0.10103535353535353)(65.0497055053711,0.10103535353535353)(65.1035304069519,0.1267929292929293)(65.12546896934509,0.15815656565656566)(65.14250755310059,0.18746632996632998)(65.14825463294983,0.21288720538720537)(65.1653995513916,0.23788720538720537)(65.17868852615356,0.2643181818181818)(65.21680569648743,0.2907996632996633)(65.36656713485718,0.31684343434343437)(65.76162981987,0.34725589225589226)(68.56196212768555,0.3722558922558923)(70.96327519416809,0.3972558922558923)(73.0741651058197,0.3972558922558923)(74.88724994659424,0.42399831649831643)(75.40257453918457,0.42399831649831643)(83.04750967025757,0.42399831649831643)(84.95266366004944,0.4511952861952862)(86.26829528808594,0.4511952861952862)(87.58007979393005,0.4525420875420876)(107.45485401153564,0.453425925925926)(122.5246069431305,0.45515993265993265)(127.13595247268677,0.4686026936026936)(127.27455163002014,0.4696717171717172)(127.33253455162048,0.4696717171717172)(127.37845396995544,0.4696717171717172)(127.38575911521912,0.49492424242424243)(127.40017032623291,0.49492424242424243)(127.42698836326599,0.49492424242424243)(127.65100169181824,0.49492424242424243)(128.09569668769836,0.49492424242424243)(130.57778096199036,0.49492424242424243)(131.85355162620544,0.49492424242424243)(132.11559343338013,0.5083249158249158)(133.87594056129456,0.5340656565656566)(136.47849893569946,0.560530303030303)(138.47949981689453,0.5699242424242424)(138.73857522010803,0.5699242424242424)(138.74832940101624,0.5699242424242424)(146.53635907173157,0.5699242424242424)(149.05072379112244,0.5727609427609428)(149.11628580093384,0.5727609427609428)(151.99667119979858,0.5727609427609428)(157.38538575172424,0.5727609427609428)(163.94331622123718,0.5727609427609428)(170.72090244293213,0.5727609427609428)(170.97172117233276,0.5727609427609428)(174.91794776916504,0.5727609427609428)(176.37658643722534,0.5727609427609428)(178.90881872177124,0.5727609427609428)(187.2972023487091,0.5727609427609428)(189.2704634666443,0.5825925925925926)(189.29636669158936,0.5825925925925926)(189.30414867401123,0.5825925925925926)(189.30567598342896,0.5832239057239057)(189.37689566612244,0.5832239057239057)(189.39843654632568,0.5832239057239057)(189.56313467025757,0.5832239057239057)(189.89261770248413,0.5861447811447812)(190.25311923027039,0.5861447811447812)(190.3247401714325,0.5861447811447812)(193.1747796535492,0.5917508417508417)(195.4705855846405,0.5917508417508417)(197.40926861763,0.5917508417508417)(197.75282406806946,0.5917508417508417)(198.50979924201965,0.592516835016835)(199.359858751297,0.592516835016835)(200.48453330993652,0.592516835016835)(200.79694294929504,0.592516835016835)(202.36247324943542,0.592516835016835)(202.3940622806549,0.5932070707070707)(207.2771120071411,0.5932070707070707)(209.97316145896912,0.5932070707070707)(212.3125638961792,0.5932070707070707)(212.62050676345825,0.5932070707070707)(214.70875668525696,0.5932070707070707)(216.9429576396942,0.5932070707070707)(218.35933423042297,0.5932070707070707)(225.01866102218628,0.5932070707070707)(225.15423560142517,0.5932070707070707)(226.54770755767822,0.5932070707070707)(226.82163524627686,0.5932070707070707)(227.4252414703369,0.5932070707070707)(228.3147406578064,0.5932070707070707)(234.17415761947632,0.594124579124579)(235.61273074150085,0.594124579124579)(242.8000156879425,0.594124579124579)(251.47182893753052,0.594124579124579)(253.07886123657227,0.5987710437710438)(253.58032488822937,0.5987710437710438)(253.69187355041504,0.6082323232323233)(253.7934935092926,0.6082323232323233)(256.8777551651001,0.6082323232323233)(262.5040547847748,0.6082323232323233)(262.7001643180847,0.6082323232323233)(263.17268800735474,0.6082323232323233)(263.43939447402954,0.6082323232323233)(264.8313250541687,0.6082323232323233)(265.25469040870667,0.6082323232323233)(266.2566683292389,0.6178872053872053)(266.76435947418213,0.6178872053872053)(272.32700872421265,0.6178872053872053)(275.7271897792816,0.6178872053872053)(277.3164839744568,0.6178872053872053)(288.37229919433594,0.6178872053872053)(288.7181158065796,0.6178872053872053)(290.7071235179901,0.6178872053872053)(291.4503126144409,0.6178872053872053)(291.5061798095703,0.6178872053872053)(291.9967429637909,0.6178872053872053)(292.25660824775696,0.6216161616161615)(298.6388840675354,0.6216161616161615)(300.80072355270386,0.6216161616161615)(300.85240387916565,0.6216161616161615)(301.15694665908813,0.6216161616161615)(305.512717962265,0.6216161616161615)(306.2417004108429,0.6216161616161615)(306.79902958869934,0.6216161616161615)(309.1679346561432,0.6216161616161615)(312.5904176235199,0.6216161616161615)(313.5332295894623,0.6216161616161615)(315.8007242679596,0.6222643097643097)(316.7063434123993,0.6222643097643097)(319.08419966697693,0.6222643097643097)(322.7501974105835,0.6222643097643097)(324.92915439605713,0.6222643097643097)(325.7642741203308,0.6222643097643097)(326.0749704837799,0.6222643097643097)(326.6383168697357,0.6222643097643097)(326.84914779663086,0.6222643097643097)(328.1142213344574,0.6272811447811447)(329.3677623271942,0.6272811447811447)(330.9665744304657,0.6272811447811447)(332.46566915512085,0.6272811447811447)(335.2599744796753,0.6272811447811447)(341.5085301399231,0.6272811447811447)(341.6103491783142,0.6272811447811447)(341.72476506233215,0.6272811447811447)(343.0452690124512,0.6272811447811447)(344.6483805179596,0.6272811447811447)(345.0371780395508,0.6272811447811447)(348.645742893219,0.6272811447811447)(350.8994629383087,0.6272811447811447)(351.85864543914795,0.6272811447811447)(352.4180965423584,0.6272811447811447)(355.02501034736633,0.6272811447811447)(355.85377526283264,0.6369360269360269)(359.02401065826416,0.6369360269360269)(359.41461539268494,0.6369360269360269)(363.71372079849243,0.6369360269360269)(365.04897689819336,0.6369360269360269)(366.6734108924866,0.6399579124579124)(369.25995206832886,0.6399579124579124)(369.2654478549957,0.6399579124579124)(374.83915281295776,0.6399579124579124)(376.3515384197235,0.6399579124579124)(376.6781256198883,0.6399579124579124)(378.5563018321991,0.6399579124579124)(380.12808299064636,0.6399579124579124)(380.5831067562103,0.6399579124579124)(382.5189175605774,0.6399579124579124)(383.69602704048157,0.6399579124579124)(384.94293189048767,0.6399579124579124)(384.9831597805023,0.6399579124579124)(385.8862977027893,0.6399579124579124)(387.2674672603607,0.6399579124579124)(388.175261259079,0.6399579124579124)(388.33831548690796,0.6399579124579124)(389.1544818878174,0.6399579124579124)(391.6340594291687,0.6399579124579124)(391.8886957168579,0.6399579124579124)(393.7944664955139,0.6399579124579124)(404.92801547050476,0.6399579124579124)(405.2780704498291,0.6399579124579124)(405.8043842315674,0.6399579124579124)(408.97934556007385,0.6399579124579124)(410.18965196609497,0.6399579124579124)(411.68111085891724,0.6399579124579124)(412.15342259407043,0.6399579124579124)(413.49223709106445,0.6399579124579124)(419.02049589157104,0.6399579124579124)(423.49045395851135,0.6399579124579124)(424.250111579895,0.6399579124579124)(426.63663959503174,0.6399579124579124)(432.6640269756317,0.6399579124579124)(433.8511929512024,0.6399579124579124)(435.4138057231903,0.6399579124579124)(440.5289959907532,0.6399579124579124)(442.37912797927856,0.6399579124579124)(445.8447289466858,0.6402272727272728)(446.81991267204285,0.6402272727272728)(447.019474029541,0.6402272727272728)(448.1075875759125,0.6438131313131313)(451.45028424263,0.6438131313131313)(451.8863687515259,0.6438131313131313)(453.92037892341614,0.6438131313131313)(457.1744952201843,0.6438131313131313)(457.20658898353577,0.6438131313131313)(457.43746423721313,0.6438131313131313)(459.3122088909149,0.6438131313131313)(459.7990794181824,0.6438131313131313)(467.2114450931549,0.6438131313131313)(470.5901141166687,0.6438131313131313)(472.374064207077,0.6438131313131313)(473.2892029285431,0.6438131313131313)(476.1393895149231,0.6438131313131313)(477.51232504844666,0.6438131313131313)(488.01125049591064,0.6458670033670033)(490.1923770904541,0.6460016835016835)(491.31475734710693,0.6460016835016835)(496.81668162345886,0.6460016835016835)(496.8619861602783,0.6460016835016835)(498.68011593818665,0.6460016835016835)(505.5616981983185,0.6460016835016835)(510.0751049518585,0.6460016835016835)(510.1912293434143,0.6460016835016835)(510.37490725517273,0.6460016835016835)(511.084171295166,0.6460016835016835)(515.2302613258362,0.646540404040404)(517.9794173240662,0.646540404040404)(518.704954624176,0.646540404040404)(519.7522878646851,0.6506144781144781)(520.6323883533478,0.6506144781144781)(521.5047543048859,0.6506144781144781)(523.2236969470978,0.6506144781144781)(523.4377653598785,0.6506144781144781)(523.6758210659027,0.6506144781144781)(523.9326786994934,0.6506144781144781)(525.4250907897949,0.6506144781144781)(530.0477635860443,0.6506144781144781)(530.2791609764099,0.6506144781144781)(531.0384666919708,0.6506144781144781)(532.1992220878601,0.6506144781144781)(534.3144111633301,0.6506144781144781)(535.1318264007568,0.6506144781144781)(535.5524406433105,0.6506144781144781)(536.6675560474396,0.6506144781144781)(537.7815170288086,0.6506144781144781)(538.4278526306152,0.6506144781144781)(538.7576167583466,0.6506144781144781)(539.7672445774078,0.6506144781144781)(540.6319515705109,0.6506144781144781)(543.3477401733398,0.6506144781144781)(543.646532535553,0.6506144781144781)(547.2447409629822,0.6506144781144781)(553.6815249919891,0.6506228956228957)(556.3583936691284,0.6506228956228957)(556.4256627559662,0.6506228956228957)(556.4710755348206,0.6506228956228957)(564.371889591217,0.6506228956228957)(569.1220757961273,0.6506228956228957)(573.4537909030914,0.6506228956228957)(573.5765171051025,0.6532659932659933)(573.7788753509521,0.6532659932659933)(578.7983882427216,0.6532659932659933)(581.6755623817444,0.6532659932659933)(585.2636620998383,0.6532659932659933)(586.4141879081726,0.6532659932659933)(589.0094783306122,0.6575084175084176)(589.1079268455505,0.6575084175084176)(590.3387341499329,0.6575084175084176)(595.1878862380981,0.6575084175084176)(595.8972475528717,0.6575084175084176)(597.8301301002502,0.6575084175084176)(598.2469482421875,0.658375420875421)(598.3810980319977,0.658375420875421)(599.2713801860809,0.658375420875421)(599.2977724075317,0.658375420875421)(599.4309797286987,0.658375420875421)(599.591052532196,0.658375420875421)(599.6908075809479,0.658375420875421)(599.7172746658325,0.658375420875421)(599.7643857002258,0.658375420875421)(599.7740271091461,0.6585016835016837)(599.7789397239685,0.6585016835016837)(599.8114466667175,0.6585016835016837)(599.8376224040985,0.6585016835016837)(599.864693403244,0.6585016835016837)(599.8912787437439,0.6585016835016837)(600.1176187992096,0.6585016835016837)(600.1268501281738,0.6585016835016837)};


\draw [color=black,line width=1mm, -stealth](30,0) -- (30,21);
\draw [color=black,line width=1mm, -stealth](30,21) -- (65,21);
\draw [color=black,line width=1mm, -stealth](65,21) -- (65,0);

\node[circle,draw] (c) at (30,21+7.5){1};
\node[circle,draw] (c) at (65+6,21){2};
\node[circle,draw] (c) at (65+6,7){3};

\end{axis}

\end{tikzpicture}

%% file: 04_evaluation.tex
\section{Experiments \label{sec:experiments}}

Our experiments aim to answer the following questions:
\begin{enumerate}
\item How does our dynamically configured AutoML system compare to state-of-the-art AutoML systems?
\item How does dynamic AutoML system configuration perform when \answerVRtwo{one or multiple} ML application constraints have been defined? 
\item Is alternating sampling more efficient than random sampling for generating the meta-learning training data?
\item How does the number of mined AutoML configurations affect the predictive performance of \system{}?
\item \answerVRone{How does a changing the hardware environment affect the predictive performance of \system{}?}
\item \answerVRone{How does the number of constraints affect meta-training?}
\end{enumerate}

\subsection{Setup}
\label{exp:setup}
We evaluate our approach on the same dataset split as used by Feurer et al.~\cite{AutoSklearn2}: $39$~meta-test datasets and $207$~meta-train datasets. To extend our framework for fairness constraints, we add $17$~fairness-related datasets provided by Neutatz et al.~\cite{DBLP:conf/sigmod/NeutatzBA21} to the meta-train datasets because common datasets do not annotate sensitive attributes that are required to measure fairness. As test datasets for fairness, we use the five fairness datasets that Ding et al. proposed to benchmark fair ML systems~\cite{ding2021retiring}.
As a prediction accuracy metric, we leverage balanced accuracy that can handle binary, multi-class, and unbalanced classification problems. 
To compare the performance across datasets, we report the average and the standard deviation across datasets by repeatedly random sampling one result out of ten runs with different seeds with replacement. This approach ensures that we report the uncertainty induced by our system and not the different hardness scales of the datasets.
Similarly, we test significance using the Mann-Whitney U rank test with $\alpha=0.05$ between repeatedly random sampled averages. We mark a number with a star~(*) if it passes this test. \answerVRthree{Note that in some cases the rounded average is very similar or the same, but one approach still passes the significance test to be better than the other approach. In these cases, we bold the results of the approach that passes the significance test.}

Due to our limited resources, we sample the meta-training dataset for two weeks, which amounts to $6,915$ meta-training instances in total. 
Further, we mine AutoML configurations for two weeks using BO for $2,000$~iterations, which amounts to $11,911$~AutoML configurations.
As AutoML parameter space, \system{} chooses (i) the hold-out fraction, which affects both the size of training and the validation set, (ii) whether to use model ensembling, (iii) whether to use incremental training, (iv) whether to reshuffle the validation split, and (v) the whole adjustable ML hyperparameter space with $302$~ML hyperparameters. Note that we do consider the time required for ensembling for the search time as it can be run in parallel to the model search as performed for Auto-Sklearn2~\cite{AutoSklearn2}.
We ran the experiments on Ubuntu 16.04 machines with 28 $\times$ \answerVRthree{Intel(R) Xeon(R) Gold 6132 CPU @ 2.60GHz} cores and 264 GB memory.

\noindent\textbf{Baselines.}
We compare our system with the state-of-the-art AutoML systems:

\newv{
\begin{enumerate}
\item \emph{TPOT}~(0.11.5) is a genetic programming-based AutoML system that optimizes feature preprocessors and ML models~\cite{TPOT}.
\item \emph{AutoGluon}~(0.3.2) is an AutoML system that focuses on model ensembling and stacking~\cite{AutoGluon}.
\item \emph{Auto-Sklearn2}~(0.14.0)~\cite{AutoSklearn2} is the latest version of the well-known AutoML system Auto-Sklearn1~\cite{feurer2015efficient} that leverages BO, meta-learning, and model ensembling to find the Sklearn~\cite{scikit-learn} ML pipelines that achieve high predictive performance. \newvv{Further, we extended the system to support the constraints for pipeline size, inference/training time, and fairness. We follow the same approach as described in Section~\ref{sec:constrainedhpo} and only add a model to the ensemble if all constraints are satisfied. This allows a fair comparison of \system{} and Auto-Sklearn2.
\item \emph{Spearmint}~\cite{spearmint} leverages BO for constrained optimization with Gaussian processes. We use the implementation by Paleyes et al.~\cite{emukit2019} and search the same ML hyperparameter space as in our static system.}
\end{enumerate}

Furthermore, we evaluate our system with and without dynamic AutoML configuration: \emph{\system{} Dynamic} and \emph{\system{} Static}:

\sloppy{
\begin{enumerate}
\item \emph{\system{} Static.} The static version covers the full ML hyperparameter space that is inspired by Auto-Sklearn1~\cite{feurer2015efficient}. It does not leverage meta-learning to optimize the search space. The details of the ML hyperparameter space are described in Section~\ref{sec:adjustableAutoML}. We use the same ML hyperparameter ranges as Auto-Sklearn1. Further, the static version always leverages hold-out validation with 33\% validation data, which is again the default validation strategy by Auto-Sklearn1. \newvv{Additionally, it always uses model ensembling and incremental training.}
\item \emph{\system{} Dynamic} implements our proposed approach. It automatically selects a subset of the full ML hyperparameter space and \newvv{identifies the hold-out validation fraction, whether to use ensembling, incremental training, and validation split reshuffling.}
\end{enumerate}
}

In the following, we focus on a comparison and insights compared to Auto-Sklearn2 since it is the most similar system compared to ours and considered as one of the strongest systems to date.

}

\begin{table*}
\centering
\caption{Search time constraint: Balanced accuracy
averaged across 10 repetitions and 39 datasets comparing \system{} to state-of-the-art AutoML systems.}
\label{tab:automlsys}
\begin{tabular}{@{}lcccccc@{}}\toprule
Strategy & 10s & 30s &1 min & 5 min & 30min & 1h  \\ \midrule
\system{} \\
\multicolumn{1}{r}{Static} & $0.43 \pm 0.02$ & $0.53 \pm 0.02$ & $0.58 \pm 0.01$ & $0.67 \pm 0.01$ & \answerVRmulti{$0.70 \pm 0.01$} & \answerVRmulti{$0.72 \pm 0.01$} \\ 
\multicolumn{1}{r}{Dynamic} & $\boldsymbol{0.57 \pm 0.01^{*}}$ & $\boldsymbol{0.67 \pm 0.01^{*}}$ & $\boldsymbol{0.70 \pm 0.01^{*}}$ & $\boldsymbol{0.74 \pm 0.00^{*}}$ & \answerVRmulti{$0.77 \pm 0.00$} & \answerVRmulti{$0.77 \pm 0.00$}\\ 
Auto-Sklearn2 opt. & $0.00 \pm 0.00$ & $0.11 \pm 0.02$ & $0.48 \pm 0.02$ & $0.74 \pm 0.02$ & \answerVRmulti{$0.80 \pm 0.00$} & \answerVRmulti{$0.81 \pm 0.00$} \\
Auto-Sklearn2 full space & $0.00 \pm 0.00$ & $0.06 \pm 0.02$ & $0.14 \pm 0.02$ & $0.70 \pm 0.03$& \answerVRmulti{$\boldsymbol{0.80 \pm 0.00^{*}}$} & \answerVRmulti{$\boldsymbol{0.81 \pm 0.00^{*}}$} \\
TPOT & $0.00 \pm 0.00$ & $0.00 \pm 0.00$ & $0.31 \pm 0.03$ & $0.47 \pm 0.04$ & \answerVRmulti{$0.67 \pm 0.02$} & \answerVRmulti{$0.68 \pm 0.01$}\\ 
AutoGluon & $0.33 \pm 0.02$ & $0.41 \pm 0.01$ &  $0.49 \pm 0.01$ & $0.62 \pm 0.01$ & \answerVRmulti{$0.77 \pm 0.01$} & \answerVRmulti{$0.79 \pm 0.00$}\\ 
Spearmint & $0.24 \pm 0.03$ & $0.36 \pm 0.03$ & $0.43 \pm 0.01$ & $0.60 \pm 0.02$ & \answerVRmulti{$0.69 \pm 0.01$} & \answerVRmulti{$0.72 \pm 0.01$}\\ 
\bottomrule 
\end{tabular}
\end{table*}


\subsection{Effectiveness on Search Time Constraints}
The most important constraint for AutoML limits the search time, which is a mandatory constraint that AutoML systems require because it is not obvious when to terminate an AutoML system. Therefore, it is crucial that our approach works well for this constraint as it also has to be fulfilled in combination with other constraints. 
We compare our dynamically configured AutoML system~\emph{\system{} Dynamic} with the same AutoML system with the default AutoML configuration~\emph{\system{} Static}. Additionally, we compare our approach to state-of-the-art AutoML systems to show the potential of our idea of constraint-driven AutoML. We note that this is the only type of constraint easily applicable to all other AutoML systems considered in this study. 

\subsubsection{Performance Comparison}
\label{subsec:experimentdiscussion}

Table~\ref{tab:automlsys} reports the average balanced accuracy for the meta-test datasets over time and across systems. 
\answerVRmulti{We focus on search times of up to 60min as most state-of-the-art AutoML systems converge in this time period.}

\answerVRmulti{First, it is noticeable that \emph{\system{}} with the default AutoML configuration outperforms TPOT~\cite{TPOT}. The reason is that \system{} leverages incremental training, which is a multi-fidelity strategy. Therefore, it can yield ML pipelines in a short time, even for large datasets.
However, \system{} with the default AutoML configuration does not outperform Auto\-Sklearn2~\cite{AutoSklearn2} and AutoGluon~\cite{AutoGluon} for larger search times. It is noteworthy that Auto-Sklearn2 is a carefully optimized version of the Auto-Sklearn system~\cite{feurer2015efficient} with a smaller hand-designed configuration space with six model classes. We also report the performance of Auto-Sklearn2 using the full ML hyperparameter space like Auto-Sklearn1. This version achieves significantly worse predictive performance, which shows that the right choice of the ML hyperparameter space is crucial.

Our approach~\system{} (Dynamic) with meta-learned AutoML configuration outperforms all other systems significantly according to the Mann-Whitney U rank test ($\alpha = 0.05$) until 5 minutes of search time. Note that both the pool of configurations that we choose the configurations from and the meta-model that chooses the configuration were generated with scenarios until 5 minutes of search time.
This finding shows that our objective of dynamically choosing good AutoML configurations was achieved if the scenarios were in the domain of the meta-training.}

In fact, \emph{\system{} Dynamic} selects on average only $55$ out of $302$~ ML hyperparameters for the search space and a 5-minute time frame and still achieves a higher average balanced accuracy across all experiments. Interestingly the search space only reduces slightly from here on. Having the 10 seconds constraint, $51$ ML hyperparameters are considered on average, which is only four less than $55$ for 5 minutes.




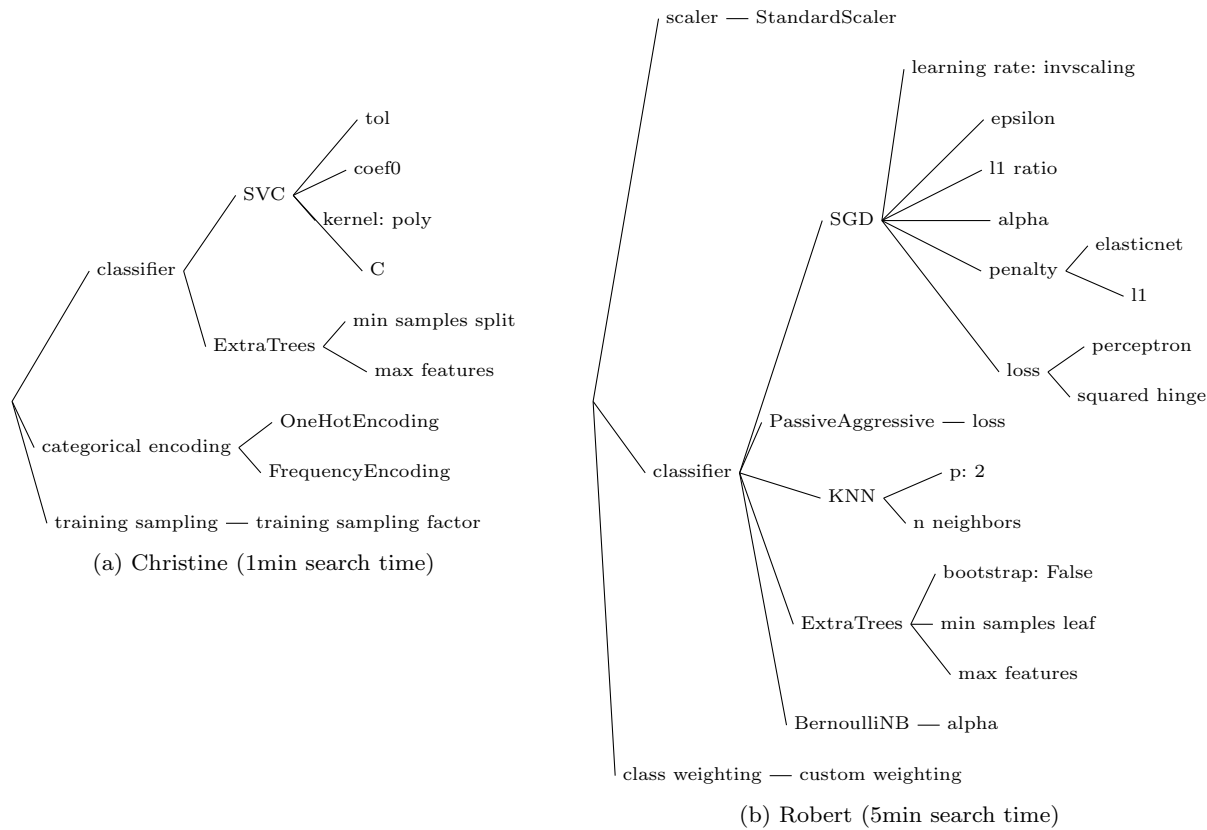
\begin{figure*}%
    \centering
    \subfloat[\centering Christine (1min search time)]{{{\scriptsize
\begin{forest} for tree={
    grow=east
}
[
	[training sampling
		[training sampling factor]
	]
	[categorical encoding
		[FrequencyEncoding]
            [OneHotEncoding]
	]
	[classifier
		[ExtraTrees
                [max features]
                [min samples split]
            ]
		[SVC
                [C]
                [kernel: poly]
                [coef0]
                [tol]
            ]
	]
]
\end{forest}
}}}%
\qquad
\subfloat[\centering Robert (5min search time)]{{{\scriptsize
\begin{forest} for tree={
    grow=east
}
[
	[class weighting
		[custom weighting]
	]
	[classifier		
            [BernoulliNB
                [alpha]
            ]
            [ExtraTrees
                [max features]
                [min samples leaf]
                [bootstrap: False]
            ]
            [KNN
                [n neighbors]
                [p: 2]
            ]
            [PassiveAggressive
                [loss]
            ]
            [SGD
                [loss
                    [squared hinge]
                    [perceptron]
                ]
                [penalty
                    [l1]
                    [elasticnet]
                ]
                [alpha]
                [l1 ratio]
                [epsilon]
                [learning rate: invscaling]
            ]
	]
        [scaler
            [StandardScaler]
        ]
]
\end{forest}
}}}%
    \caption{Examples of ML hyperparameter spaces chosen by \system{} Dynamic}%
    \label{fig:example_spaces}%
\end{figure*}

\newvv{Yet, the space can also differ significantly between $5$ and $1$ minutes. Figure~\ref{fig:example_spaces} shows AutoML configurations that were selected for the dataset ``Christine'' and ``Robert''  from the OpenML repository. The visualization follows the hierarchical view that we presented in Section~\ref{sec:adjustableAutoML} and displays the obtained configuration space for 1min and 5min search time, respectively. 
Comparing the ML hyperparameter spaces, we see that in this case the ML hyperparameter space for 1min search time is smaller than for 5min search time. This is because a higher time period allows for the optimization of more ML pipeline parameters. 

Additionally, for the dataset ``Christine'', our system chooses the validation fraction~$0.13$, ensembling, and incremental training. The small validation fraction reduces the time for evaluation. Ensembling makes the predictions more robust against noise. Incremental training ensures that the system finds a suitable ML pipeline early. In addition to incremental training, our system also chose to optimize the size of training set to further reduce the iteration overhead.

For the dataset ``Robert'', our system chooses the validation fraction~$0.54$, incremental training, and validation split reshuffling.
Validation split reshuffling avoids overfitting. Additionally, our system chose to optimize each class weight individually because the dataset has 10 classes.

\begin{table*}
\centering
\caption{\answerVRone{AutoML parameters chosen by \system{} Dynamic for different search times on the dataset ``numerai28.6''.}}
\label{tab:spacesonedata}
\begin{tabular}{@{}ccc@{}}\toprule
10s & 1min & $\geq$ 5min \\ \midrule
\input{10s} & \input{1min} & \input{5min}\\ \midrule
holdout: 0.45 & holdout: 0.61 & holdout: 0.46 \\
ensemble: No & ensemble: Yes & ensemble: No \\
random shuffle: Yes & random shuffle: No & random shuffle: Yes \\
incremental: Yes & incremental: Yes & incremental: No \\
\bottomrule 
\end{tabular}
\end{table*}

\answerVRone{Table~\ref{tab:spacesonedata} presents an example that shows the AutoML parameters chosen for the dataset ``numerai28.6'' under different search time constraints: 10s, 1min, and $\geq$ 5min.
Since our \system{} Dynamic was trained on the data until 5 minutes, it will pick the same search space for search times greater than 5 minutes, which is why we did not consider higher search time constraints here.

Even for the very short search time of 10s and the rather large dataset with 96,320 instances, the search space still incorporates 9 out of 16 classifiers because of the incremental training, which enables fast skipping of poorly performing ML pipelines. 
For 1min, our system increases both the search space and the holdout fraction. With this change, the holdout validation score evaluation will take more time but will be more accurate.
For $\geq$ 5min, our system chooses to avoid incremental training. This way, model training will take more time but the models will be trained on more instances and are more likely to achieve a better generalization.
}

}

\begin{table}
\centering
\caption{Meta-feature importances of the classification meta-model.}
\label{tab:featureimportance}
\begin{tabular}{@{}rlc@{}}\toprule
Rank & Meta-Feature & Importance \\ \midrule
1 & pipeline size constraint & 0.072 \\ 
2 & inference time constraint & 0.053 \\ 
3 & training time constraint & 0.044 \\ 
4 & hold-out fraction & 0.036 \\ 
5 & search time constraint & 0.023 \\ 
6 & number of evaluations & 0.022 \\ 
7 & fairness constraint & 0.020 \\ 
8 & hold-out test instances & 0.017 \\ 
9 & evaluation time & 0.017 \\ 
10 & $|\text{instances}|$ & 0.016 \\ 
11 & ClassProbabilitySTD & 0.016 \\ 
12 & DatasetRatio & 0.015 \\ 
13 & ClassProbabilityMax & 0.015 \\ 
14 & ClassProbabilityMin & 0.015 \\ 
15 & ClassEntropy & 0.015 \\
\bottomrule 
\end{tabular}
\end{table}

\subsubsection{Analyzing the Meta-Models}
\label{sub:analysis}
\begin{sloppypar}
To analyze the meta-models, we computed the meta-feature importance based on impurity scores for the trained random forest meta-model. We list the top-15 meta-features in Table~\ref{tab:featureimportance} for the classification meta-model. 
The most important meta-features are the constraint thresholds, in particular, for the pipeline size, and inference/training time.
These meta-features are important because different constraints also require different AutoML configurations. This finding supports the aim of this work to consider dynamic AutoML configuration, especially for constrained settings.
Another important feature is the hold-out fraction. Especially for large datasets, it is crucial to identify the right sample size to allow the AutoML system to yield any ML pipeline. 
For instance, for the dataset ``KDDCup09 appetency'' ($50$k~instances), our method chooses a validation fraction of $7\%$ of the data.

The remaining 8th-15th meta-features all cover dataset-specific meta-features, e.g. about the class distributions and the shape of the data. The meta-features representing the ML hyperparameter search space are less important, e.g. the meta-feature of whether to use a specific categorical encoding is the 37th most important feature.
\end{sloppypar}

\begin{table}
\centering
\caption{Meta-feature importances of the regression meta-model.}
\label{tab:featureimportance_regression}
\begin{tabular}{@{}rlc@{}}\toprule
Rank & Meta-Feature & Importance \\ \midrule
1 & pipeline size constraint & 0.072 \\
2 & inference time constraint & 0.056 \\
3 & training time constraint & 0.052 \\
4 & hold-out fraction & 0.043 \\
5 & search time constraint & 0.024 \\
6 & preprocessor & 0.023 \\
7 & fairness constraint & 0.022 \\
8 & number of evaluations & 0.022 \\
9 & incremental training & 0.021 \\
10 & ClassProbabilitySTD & 0.020 \\
11 & ClassProbabilityMin & 0.016 \\
12 & ClassEntropy & 0.016 \\
13 & ClassProbabilityMax & 0.015 \\
14 & evaluation time & 0.015 \\
15 & RatioNominalToNumerical & 0.015 \\
\bottomrule 
\end{tabular}
\end{table}

For the regression meta-model, we list the top-15 meta-features in Table~\ref{tab:featureimportance_regression}. The most important meta-features are similar to the ones for the classification meta-model. However, for the regression meta-model, the meta-feature that describes whether to use a feature \emph{preprocessor} and whether to \emph{incremental training}. Both decisions have a significant impact on how much the given AutoML configuration outperforms the default one.

\begin{table*}
\smallestfont
\centering
\caption{Choice of the classifiers: AdaBoost~(AdaB.), 
Bernoulli Naive Bayes~(B.NB), 
Decision Tree~(DT),
Extra Trees~(E.Trees),
Gaussian Naive Bayes~(G.NB),
Histogram-based Gradient Boosting~(HGB.),
K-Nearest Neighbors~(KNN),
Linear Discriminant Analysis~(LDA),
Linear Support Vector Classification~(LSVC),
Multi-layer Perceptron~(MLP), 
Multinomial Naive Bayes~(M.NB),
Passive Aggressive~(PA),
Quadratic Discriminant Analysis~(QDA),
Random Forest~(RF),
Stochastic Gradient Descent~(SGD),
Support Vector Classification~(SVC).
}
\label{tab:classifiers}
\begin{tabular}{@{}lccccccccccccccccc@{}}\toprule
Time & AdaB. & B.NB & DT & E.Trees & G.NB & HGB. & KNN & LDA & LSVC & MLP & M.NB & PA & QDA & RF & SGD & SVC & $|\text{clf.}|$ \\ \midrule
10s   & 0.69 & 0.79 & 0.54 & 0.90 & 0.51 & 0.31 & 0.62 & 0.59 & 0.54 & 0.49 & 0.44 & 0.51 & 0.67 & 0.31 & 0.69 & 0.74 & 9.33\\ 
30s   & 0.82 & 0.77 & 0.51 & 0.95 & 0.67 & 0.28 & 0.62 & 0.69 & 0.49 & 0.46 & 0.54 & 0.56 & 0.69 & 0.28 & 0.67 & 0.90 & 9.90\\ 
1min  & 0.85 & 0.74 & 0.51 & 0.97 & 0.62 & 0.28 & 0.69 & 0.72 & 0.41 & 0.44 & 0.56 & 0.41 & 0.72 & 0.28 & 0.72 & 0.87 & 9.79\\ 
5min  & 0.79 & 0.82 & 0.72 & 0.95 & 0.59 & 0.38 & 0.72 & 0.54 & 0.54 & 0.49 & 0.54 & 0.64 & 0.72 & 0.36 & 0.82 & 0.85 & 10.46\\
\answerVRone{30min} & \answerVRone{0.79} & \answerVRone{0.82} & \answerVRone{0.72} & \answerVRone{0.95} & \answerVRone{0.59} & \answerVRone{0.38} & \answerVRone{0.72} & \answerVRone{0.54} & \answerVRone{0.54} & \answerVRone{0.49} & \answerVRone{0.54} & \answerVRone{0.64} & \answerVRone{0.72} & \answerVRone{0.36} & \answerVRone{0.82} & \answerVRone{0.85} & \answerVRone{10.46}\\ 
\answerVRone{1h}    & \answerVRone{0.79} & \answerVRone{0.82} & \answerVRone{0.72} & \answerVRone{0.95} & \answerVRone{0.59} & \answerVRone{0.38} & \answerVRone{0.72} & \answerVRone{0.54} & \answerVRone{0.54} & \answerVRone{0.49} & \answerVRone{0.54} & \answerVRone{0.64} & \answerVRone{0.72} & \answerVRone{0.36} & \answerVRone{0.82} & \answerVRone{0.85} & \answerVRone{10.46}\\ 

\bottomrule 
\end{tabular}
\end{table*}

Table~\ref{tab:classifiers} contains statistics about how often our system chooses a specific classifier across the 39 datasets and how many classifiers it chooses on average.
The first observation is that the meta-model learned that it is beneficial to choose around ten~classifiers to achieve high balanced accuracy fast. The Auto-Sklearn2 developers choose only 5 classifiers. However, since our system can decide for every single ML hyperparameter whether to optimize it, the search space stays small in comparison but adjusts itself to the specified dataset. In contrast to building Auto-Sklearn2, this approach is fully automatic and does not require any AutoML systems expertise. \newv{Auto-Sklearn2 uses a dynamic chooses the validation strategy. Additionally, its ML hyperparameter space has been manually tuned for accuracy and search time. Thus, users who want to apply Auto-Sklearn2 for a new constrained setting, would need to adjust the ML hyperparameter search space manually again.}
Further, we see that ExtraTrees are chosen frequently. The reason is that the computation cost is low and the prediction is robust due to ensembling.

\answerVRone{For some classifiers, such as MLP and HGB, the frequency stays similar across search time constraints. The reason is twofold: First, using incremental training, we can quickly yield working models for both classifier types that are competitive across search times. Second, \system{} identified that these classifier types work well for specific datasets which do not change across constraints. For instance, HGB was chosen for balanced datasets with less than 8 classes and more than 57 numeric features. MLP was chosen for skewed datasets with many categorical features.

Further, for some classifiers, such as LDA and SVC, the frequency increases with increasing search time and then decreases again. For instance, LDA benefits from an increasing number of training instances but is prone to overfitting for unbalanced data if one optimizes it for long enough. The training of SVC is very efficient and therefore, one can train an SVC with many instances in very little time. Therefore, we see a high frequency of 90\% for 30s. With increasing search time, other more complex models, such as RF, replace it incrementally. 

Finally, the frequency across models stays the same because both the pool of configurations that we sample from was generated with a maximum search time of 5mins and training data of the model that chooses the configuration has the same limit.
}

\begin{table}
\centering
\caption{Choice of AutoML parameters.}
\label{tab:otherparams}
\begin{tabular}{lcccc} \toprule
Search Time & 10s & 30s & 1min & 5min \\  \midrule
Incremental training & 0.97 & 0.97 & 0.90 & 0.82 \\
Ensemble & 0.57 & 0.55 & 0.56 & 0.62 \\
Class augmentation & 0.37 & 0.24 & 0.21 & 0.10 \\
Validation split reshuffling & 0.17 & 0.29 & 0.26 & 0.26 \\ \bottomrule
\end{tabular}
\end{table}

To understand the interaction among the other AutoML parameters, we report in Table~\ref{tab:otherparams} the fraction of datasets that a certain AutoML parameter was applied. First, we see that the choice for incremental training in most cases only decreases slightly with increasing search time. Incremental training ensures that we find ML pipelines independent of the dataset size. Model ensembling is also used frequently because it ensures robustness. Class augmentation is not applied often because most datasets are already balanced. Additionally, its use decreases with increasing search time. The reason might be that with a long enough search time, we can find a suitable model that internally addresses the class imbalance.
Finally, validation split reshuffling is increasingly used with increasing search time. Greater search times lead to a higher number of iterations that in turn raise the risk of overfitting and reshuffling can help to reduce this risk. 
To the best of our knowledge, none of the state-of-the-art systems leverage this reshuffling strategy. Our results show that it is promising and justifies further research.


The choice of feature preprocessors is on par with the Auto-Sklearn2 implementation. Auto-Sklearn2 does not perform any feature preprocessing, and our dynamic approach follows the same strategy. 
The reason is that feature preprocessing might add more overhead than benefit for the predictive performance.

\subsubsection{Conclusion}
Our simple AutoML system with the default configuration is already competitive to state-of-the-art systems, such as AutoGluon and TPOT. This might be due to the fact that our approach leverages incremental training, therefore, can handle large datasets.
Second, our dynamic AutoML configuration approach outperforms the same system with the default configuration for all search time constraints. Third, our dynamic approach is head-to-head with the hand-tuned Auto-Sklearn2 system, which was tuned by (Auto)ML system experts.

\begin{table*}
\centering
\caption{We report the balanced accuracy for 5 minutes search time averaged across 10 repetitions and test datasets for four constraints.}
\label{tab:constraints}
\begin{tabular}{@{}lccccc@{}}\toprule
Percentile & 2\% & 4\% & 8\% & 16\% & 32\% \\ \midrule
Pipeline & \multirow{2}{*}{4026B} & \multirow{2}{*}{6651B} & \multirow{2}{*}{8359B} & \multirow{2}{*}{16797B} & \multirow{2}{*}{32266B} \\
Size \\ \midrule
Auto-Sklearn2 & $ 0.01 \pm 0.00 $ & $ 0.01 \pm 0.00 $ & $ 0.01 \pm 0.00 $ & $ 0.01 \pm 0.00 $ & $ 0.01 \pm 0.00$ \\
Spearmint & $ 0.04 \pm 0.02 $ & $ 0.08 \pm 0.02 $ & $ 0.09 \pm 0.03 $ & $ 0.19 \pm 0.03 $ & $ 0.22 \pm 0.03$ \\
\system{} \\
\multicolumn{1}{r}{Dynamic} & $ 0.25 \pm 0.01 $ & $ \boldsymbol{0.39 \pm 0.01^{*}} $ & $ \boldsymbol{0.43 \pm 0.00^{*}} $ & $ \boldsymbol{0.54 \pm 0.01^{*}} $ & $ \boldsymbol{0.63 \pm 0.01^{*}}$ \\
\multicolumn{1}{r}{Static} & $ \boldsymbol{0.25 \pm 0.01^{*}} $ & $ 0.39 \pm 0.01 $ & $ 0.42 \pm 0.01 $ & $ 0.49 \pm 0.01 $ & $ 0.59 \pm 0.01$ \\
\bottomrule
Training & \multirow{2}{*}{0.009s} & \multirow{2}{*}{0.010s} & \multirow{2}{*}{0.012s} & \multirow{2}{*}{0.019s} & \multirow{2}{*}{0.078s} \\
time \\   \midrule
Auto-Sklearn2  & $ 0.00 \pm 0.00 $ & $ 0.00 \pm 0.00 $ & $ 0.00 \pm 0.00 $ & $ 0.00 \pm 0.00 $ & $ 0.01 \pm 0.01$ \\
Spearmint  & $ 0.00 \pm 0.01 $ & $ 0.01 \pm 0.01 $ & $ 0.00 \pm 0.01 $ & $ 0.00 \pm 0.01 $ & $ 0.05 \pm 0.02$ \\
\system{} \\
\multicolumn{1}{r}{Dynamic} & $ \boldsymbol{0.61 \pm 0.01^{*}} $ & $ \boldsymbol{0.62 \pm 0.01^{*}} $ & $ \boldsymbol{0.63 \pm 0.01^{*}} $ & $ \boldsymbol{0.68 \pm 0.01^{*}} $ & $ \boldsymbol{0.71 \pm 0.00^{*}}$ \\
\multicolumn{1}{r}{Static} & $ 0.46 \pm 0.02 $ & $ 0.46 \pm 0.02 $ & $ 0.50 \pm 0.02 $ & $ 0.57 \pm 0.02 $ & $ 0.65 \pm 0.01$ \\
\bottomrule 
Inference & \multirow{2}{*}{0.00079s} & \multirow{2}{*}{0.00082s} & \multirow{2}{*}{0.00102s} & \multirow{2}{*}{0.00146s} & \multirow{2}{*}{0.00302s} \\
time \\ \midrule
Auto-Sklearn2 & $ 0.29 \pm 0.02 $ & $ 0.27 \pm 0.02 $ & $ 0.27 \pm 0.03 $ & $ 0.40 \pm 0.02 $ & $ 0.42 \pm 0.02$ \\
Spearmint & $ 0.02 \pm 0.01 $ & $ 0.02 \pm 0.02 $ & $ 0.02 \pm 0.01 $ & $ 0.02 \pm 0.01 $ & $ 0.06 \pm 0.01$ \\
\system{} \\
\multicolumn{1}{r}{Dynamic} & $ \boldsymbol{0.42 \pm 0.02^{*}} $ & $ \boldsymbol{0.45 \pm 0.02^{*}} $ & $ \boldsymbol{0.57 \pm 0.02^{*}} $ & $ \boldsymbol{0.66 \pm 0.01^{*}} $ & $ \boldsymbol{0.74 \pm 0.00^{*}}$ \\
\multicolumn{1}{r}{Static}  & $ 0.25 \pm 0.03 $ & $ 0.26 \pm 0.03 $ & $ 0.38 \pm 0.03 $ & $ 0.52 \pm 0.02 $ & $ 0.64 \pm 0.02$ \\
\bottomrule 
Equal & \multirow{2}{*}{1.000} & \multirow{2}{*}{0.999} & \multirow{2}{*}{0.994} & \multirow{2}{*}{0.981} & \multirow{2}{*}{0.949} \\
Opportunity \\ \midrule
Auto-Sklearn2 & $ \boldsymbol{0.50 \pm 0.00^{*}} $ & $ 0.56 \pm 0.00 $ & $ 0.59 \pm 0.01 $ & $ 0.63 \pm 0.00 $ & $ 0.67 \pm 0.02$ \\
Spearmint & $ 0.17 \pm 0.10 $ & $ 0.19 \pm 0.12 $ & $ 0.35 \pm 0.11 $ & $ 0.57 \pm 0.07 $ & $ 0.58 \pm 0.07$ \\
\system{} \\
\multicolumn{1}{r}{Dynamic} & $ 0.10 \pm 0.00 $ & $ \boldsymbol{0.61 \pm 0.05^{*}} $ & $ \boldsymbol{0.64 \pm 0.01^{*}} $ & $ \boldsymbol{0.67 \pm 0.01^{*}} $ & $ \boldsymbol{0.70 \pm 0.00^{*}}$ \\
\multicolumn{1}{r}{Static} & $ 0.10 \pm 0.00 $ & $ 0.46 \pm 0.09 $ & $ 0.62 \pm 0.05 $ & $ 0.66 \pm 0.01 $ & $ 0.68 \pm 0.01$ \\
\bottomrule 
\end{tabular}
\end{table*}

\subsection{Effectiveness on Diverse Constraint Types}
\label{sec:oneconstrainteval}
To evaluate that our approach also achieves high balanced accuracy for constraint types other than search time, we provide experiments with constraints on ML pipeline size, the training time, the inference time, and equal opportunity~(fairness metric).
For the experimental setting, we obtain constraint thresholds for the different constraint types in the following way. We ran random ML tasks for one day and obtained the distributions across all evaluated ML pipelines for all constraint types. On these, we can compute different percentiles to simulate different tight constraints.
For training the meta-model, \emph{\system{} Dynamic} could freely choose any of the percentiles (with a maximum search time of 5 minutes). To compare \system{} to the baselines, we used fairly tight constraints, i.e. the 2nd, 4th, 8th, 16th, and 32nd percentile of each distribution. We also evaluated the 1st percentile but the results are similar to the 2nd percentile and due to space limitations, we omit the corresponding results.

\begin{sloppypar}
Other state-of-the-art systems, such as AutoGluon, TPOT, and Auto-Sklearn2, do not natively support these ML application constraints and are not easily extensible because their API only provides access to the ML pipeline predictions. \newvv{However, we extended the best-performing AutoML system Auto-Sklearn2 to support constraints to allow a comparison to our system. 
Furthermore, constrained BO systems, such as Spearmint~\cite{spearmint}, GPflowOpt~\cite{gpflowopt}, \newv{ADMMBO~\cite{DBLP:journals/jmlr/AriafarCBD19}}, and Ax~\cite{Ax}, support arbitrary constraints for BO.
As a representative of this class of systems, we benchmark Spearmint with the ML hyperparameter space of our static system.}
\end{sloppypar}

\subsubsection{Performance Comparison}
\begin{sloppypar}
Table~\ref{tab:constraints} provides the results of these constraint thresholds. 
%
\newvv{Across constraint types, \system{} outperforms both baselines significantly. Only for equal opportunity, Auto-Sklearn2 achieves the best accuracy for very restrictive fairness constraints. The reason is that Auto-Sklearn2 uses Dummy classifiers if it does not find any other model. Dummy classifiers predict only one class. This way it is likely that both the majority and the minority group have very similar true positive rates and therefore very high equal opportunity. However, we decided against including dummy classifiers because users expect an AutoML system to fit actual ML models.

For the constraints inference and training time, our dynamic approach always outperforms our static approach. For pipeline size constraints, the static approach is better for restrictive thresholds. The reason is that pipeline size is more bound to the size of training set size and our default approach always uses incremental training. That means that it starts with a very small training dataset. So, if the pipeline size is not satisfied for such a small set, it will go to the next ML hyperparameter configuration immediately. Our meta-model might be too optimistic and try to avoid incremental training if possible because it has a higher chance of higher accuracies but might miss satisfying the constraints.

For fairness constraints, the dynamic and static approach perform similarly. The reason is that fairness is highly data dependent. Without explicit information about the sensitive attributes, it is harder for the meta-model to decide on the AutoML system configuration. Furthermore, the meta-training for fairness had access to much fewer datasets compared to the other constraints. Additional datasets might help the meta-model to generalize better. However, in case of missing values and fairness constraints, \system{} independently learned to choose only median value imputation, which supports the finding by Schelter et al.~\cite{FairPrep} that mean value imputation negatively affects fairness. 
}

\end{sloppypar}

\begin{table*}
\footnotesize
\centering
\caption{We report the average percentage difference in choice of ML classifiers depending on the ML application constraint.}
\label{tab:classifiers_constraints}
\setlength{\tabcolsep}{1.5pt}
\begin{tabular}{@{}lccccccccccccccccc@{}}\toprule
ML Application & \multirow{2}{*}{AdaB.} & \multirow{2}{*}{B.NB} & \multirow{2}{*}{DT} & \multirow{2}{*}{E.Trees} & \multirow{2}{*}{G.NB} & \multirow{2}{*}{HGB.} & \multirow{2}{*}{KNN} & \multirow{2}{*}{LDA} & \multirow{2}{*}{LSVC} & \multirow{2}{*}{MLP} & \multirow{2}{*}{M.NB} & \multirow{2}{*}{PA} & \multirow{2}{*}{QDA} & \multirow{2}{*}{RF} & \multirow{2}{*}{SGD} & \multirow{2}{*}{SVC} \\
Constraint\\ \midrule
None & 0.79 &  0.82 & 0.72 & 0.95 & 0.59 & 0.38 & 0.72 & 0.54 & 0.54 & 0.49 & 0.54 & 0.64 & 0.72 & 0.36 & 0.82 & 0.85 \\ \midrule
Pipeline Size & \collm{-0.19} & \coll{-0.43} & \collm{-0.15} & \coll{-0.38} & -0.07 & +0.06 & \collm{-0.19} & +0.03 & -0.01 & \coll{-0.20} & \coll{-0.25} & \coll{-0.25} & \collm{-0.14} & -0.03 & \coll{-0.24} & \coll{-0.30} \\
Training Time  & +0.07 & -0.04 & +0.03 & +0.00 & \collm{-0.17} & +0.00 & +0.03 & \cole{+0.26} & \cole{+0.21} & -0.09 & +0.05 & \colem{+0.12} & +0.03 & +0.02 & +0.08 & +0.09 \\
Inference Time & \collm{-0.12} & +0.03 & -0.01 & -0.09 & \colem{+0.14} & \colem{+0.12} & \collm{-0.13} & \colem{+0.16} & \colem{+0.14} & \cole{+0.20} & \colem{+0.16} & +0.06 & +0.00 & \cole{+0.40} & \collm{-0.19} & -0.01 \\
Equal Opp.      & \coll{-0.23} & -0.02 & \colem{+0.12} & \collm{-0.11} & \coll{-0.23} & \coll{-0.22} & \coll{-0.52} & +0.06 & \colem{+0.14} & \colem{+0.19} & \coll{-0.26} & \coll{-0.24} & \collm{-0.12} & \collm{-0.16} & \collm{-0.10} & -0.01 \\
\bottomrule 
\end{tabular}
\end{table*}

\subsubsection{Analysis}
\begin{sloppypar}
To better understand how our system adapts the ML hyperparameter search space depending on the ML application constraints, we average the chosen classifiers for each ML application constraint and compare it to case using with no ML application constraint in Table~\ref{tab:classifiers_constraints}.

For the pipeline size constraint, \system{} avoids models that require more memory, such as extra trees~(E.Trees), multi-layer perception~(MLP), or the KNN classifier, which needs to store all training instances for inference.
For the training time constraint, \system{} shifts more to linear models, such as Linear Discriminant Analysis~(LDA), Linear Support Vector Classification~(LSVC), or Passive Aggressive~(PA), because they can be trained faster.
For the inference time constraint, \system{} chooses significantly more often random forest to be part of the search space because its inference complexity is only $\mathcal{O}(t\log{}n)$ where $t$ is the number of trees and $n$ is the number of instances.
For equal opportunity, \system{} avoids models that amplify the bias in the data. For instance, KNN might amplify bias because it always decides based on the majority of the nearest neighbors.
These insights confirm that we cannot optimize an AutoML system for one constraint and expect that the same optimization will also benefit other constraints.
\end{sloppypar}

\begin{figure}
	\centering
	\includegraphics[scale=0.15]{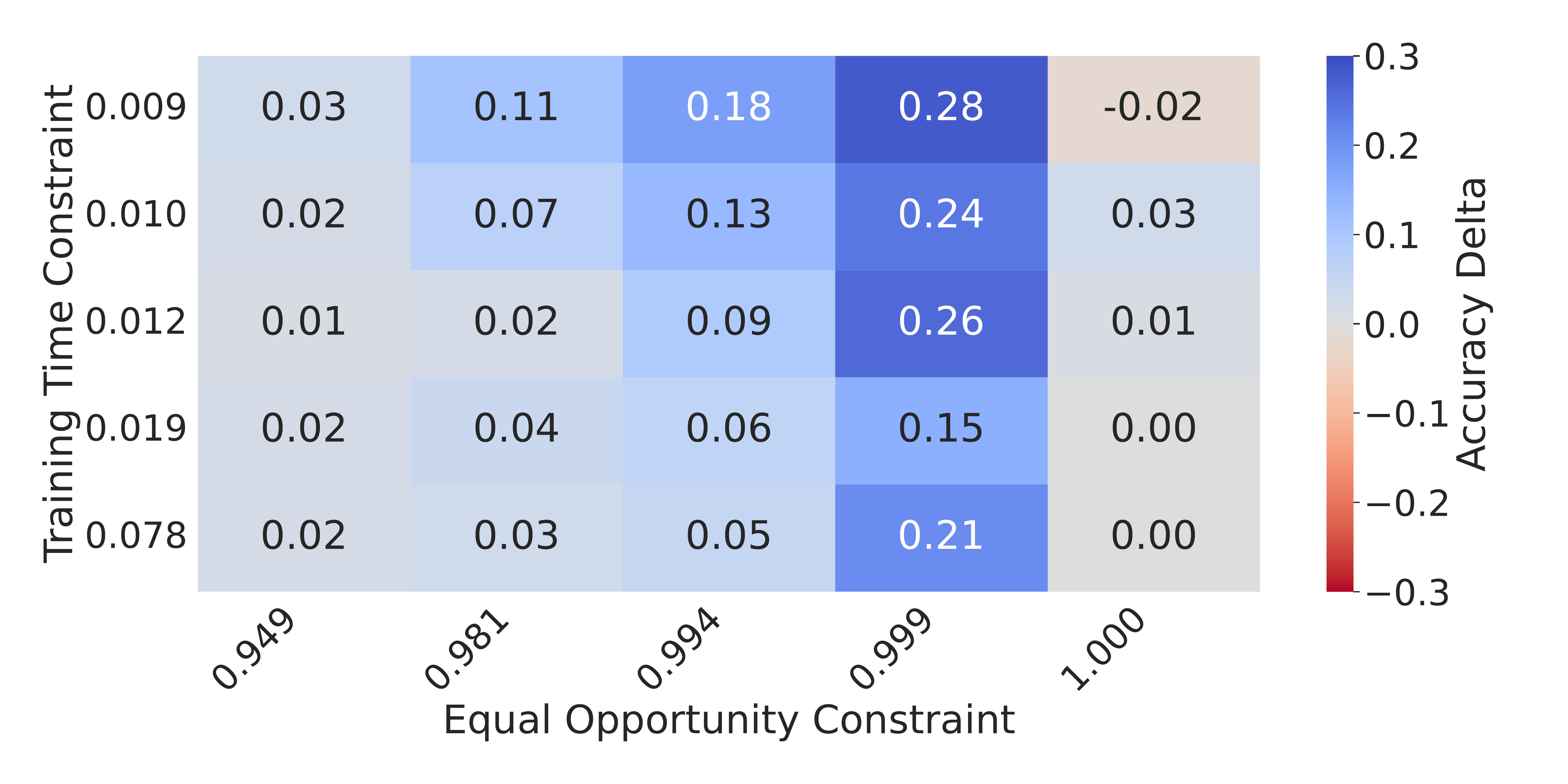}
	\caption{\answerVRtwo{We apply the constraints training time and fairness simultaneously and report the absolute distance to the average performance between the static and dynamic \system{}. Higher numbers are better.}}
	\label{figure:twoconstraints}
	\vspace{-1.0em}
\end{figure}

\begin{figure}
	\centering
	\includegraphics[scale=0.15]{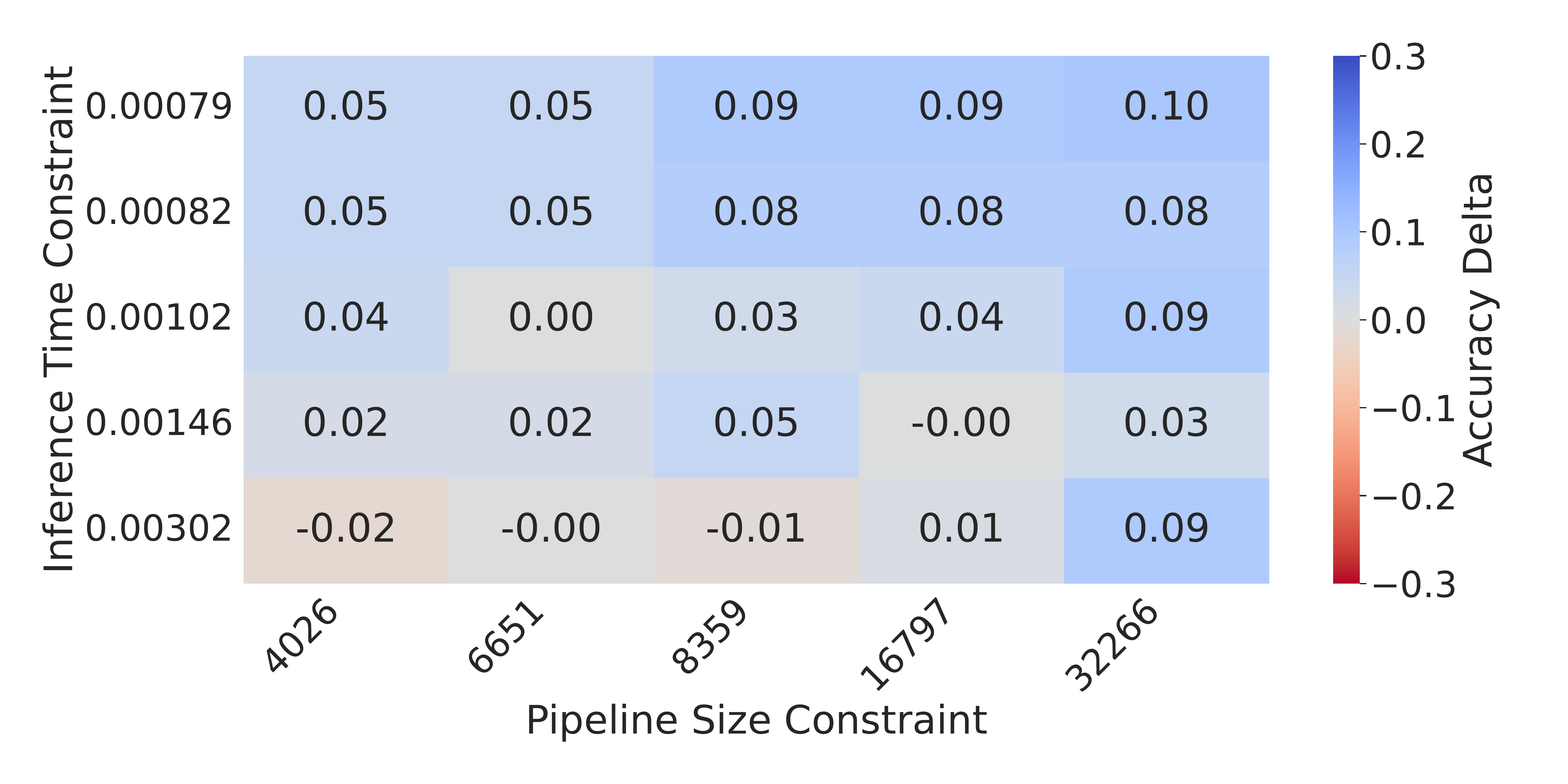}
	\caption{\answerVRtwo{We apply the constraints inference time and pipeline size simultaneously and report the absolute distance to the average performance between the static and dynamic \system{}. Higher numbers are better.}}
	\label{figure:twoconstraints2}
	\vspace{-1.0em}
\end{figure}

\answerVRtwo{
\subsection{Effectiveness on Multiple Constraint Types}
\label{sec:twoconstrainteval}
In practice, ML applications can be constrained in multiple dimensions simultaneously. 
To evaluate our system for multiple constraints simultaneously, we choose two constraint combinations training time/equal opportunity and inference time/pipeline size. For both constraint combinations, we apply all combinations of thresholds that were evaluated in Section~\ref{sec:oneconstrainteval}. We report the difference in the average balanced accuracy that \emph{\system{} Dynamic} outperforms the static variant in Figures~\ref{figure:twoconstraints} and~\ref{figure:twoconstraints2}.

In nearly all experiments, \emph{\system{} Dynamic} outperforms the static variant or achieves similar predictive performance. Only for very restrictive constraints, such as 100\% equal opportunity or 4026B pipeline size, its performance was slightly lower because these constraints were for some of the test datasets not satisfiable. Overall, the experiments show that \emph{\system{} Dynamic} even works for multiple constraints. This finding shows that our meta-learning approach learns how these different constraints interact with each other. In 65\% of the cases, \system{} chooses AutoML configurations that consider two constraints simultaneously and were never chosen for the cases where we enforced only one of the constraints.
}

\begin{table}
\centering
\caption{Predictive performance over meta-training time. We report the average balanced accuracy over training time averaged across 10 repetitions and 39 datasets comparing our system with random and alternating sampling.}
\label{tab:automl_surrogate_training}
\begin{tabular}{@{}lcc@{}}\toprule
Days & Alternating & Random\\ \midrule
2 & $\boldsymbol{0.70 \pm 0.01^{*}}$ & $0.67 \pm 0.01$\\
4 & $\boldsymbol{0.72 \pm 0.01^{*}}$ & $0.65 \pm 0.02$\\
6 & $0.70 \pm 0.01$ & $\boldsymbol{0.71 \pm 0.01^{*}}$\\
8 & $\boldsymbol{0.72 \pm 0.01^{*}}$ & $0.71 \pm 0.02$\\
10 & $\boldsymbol{0.73 \pm 0.01^{*}}$ & $0.72 \pm 0.01$\\
12 & $\boldsymbol{0.73 \pm 0.01^{*}}$ & $0.69 \pm 0.01$\\
14 & $\boldsymbol{0.74 \pm 0.00^{*}}$ & $0.72 \pm 0.01$\\ 
\bottomrule 
\end{tabular}
\end{table}

\subsection{Alternating vs Random Sampling}
One major design decision of our system is to leverage active learning in addition to random sampling to explore the huge space of AutoML parameters and constraints more efficiently. 
Table~\ref{tab:automl_surrogate_training} provides the balanced accuracy for meta-models for both sampling approaches across two weeks of training data generation. 

The alternating sampling approach outperforms the random sampling \emph{significantly}. The reason is that active learning ensures that we sample along the decision boundary while random sampling ensures the diversity in the training data. Following a purely random sampling strategy results in lower final prediction performance and less consistent gains. For instance, after $12$~days of random sampling, we achieve a worse predictive performance than for six days of sampling.

We can leverage Table~\ref{tab:automl_surrogate_training} also to understand the impact of the training time. The numbers show that more training time benefits the meta-model, and even on the 14th day, we gain 1\% more in average balanced accuracy. 
To conclude, alternating sampling outperforms random sampling significantly, and the longer we train, the better the dynamic AutoML configuration works.

\begin{table}
\centering
\caption{Predictive performance over different numbers of mined AutoML configurations for search time of 5min. We report the average balanced accuracy over training time averaged across 10 repetitions and 39 datasets.}
\label{tab:number_configs}
\begin{tabular}{@{}clc@{}}\toprule
Fraction & \# Configurations & Accuracy\\ \midrule
0.0002 & 3     & $0.729 \pm 0.00$ \\
0.0005 & 6     & $0.734 \pm 0.00$ \\
0.0010 & 12    & $0.736 \pm 0.00$ \\
0.0020 & 23    & $0.732 \pm 0.00$ \\
0.0039 & 47    & $0.736 \pm 0.00$ \\
0.0078 & 93    & $0.734 \pm 0.01$ \\
0.0156 & 186   & $0.739 \pm 0.01$ \\
0.0313 & 372   & $0.739 \pm 0.01$ \\
0.0625 & 744   & $0.739 \pm 0.01$ \\
0.1250 & 1489  & $0.735 \pm 0.01$ \\
0.2500 & 2978  & $0.744 \pm 0.00$ \\
0.5000 & 5956  & $0.743 \pm 0.00$ \\
1.0000 & 11911 & $0.747 \pm 0.00$ \\
\bottomrule 
\end{tabular}
\end{table}

\subsection{AutoML Configuration Mining}
Another important question for our approach is how many promising AutoML configurations we need to mine to achieve high predictive performance. Therefore, we experiment, for the search time constraint of 5min, with various fractions of the AutoML configurations that we mined within two weeks. We report the results in Table~\ref{tab:number_configs}. With an increasing number of mined AutoML configurations, the predictive performance increases as well. The accuracy gain in percent might seem small but it is significant according to the Mann-Whitney U rank test. Further, the more constraints we add, the more diverse the pool of mined AutoML configurations needs to be to achieve high predictive performance across all constraints.

\begin{table*}
\centering
\caption{\answerVRone{Search time constraint: Balanced accuracy
averaged over 10 repetitions and 39 datasets comparing \system{} to hardware adjusted \system{}.}}
\label{tab:benchres}
\begin{tabular}{@{}lccccc@{}}\toprule
\system{} Strategy & 10s & 30s &1 min & 5 min   \\ \midrule
Static & $0.59 \pm 0.01$ & $0.66 \pm 0.01$ & $0.68 \pm 0.01$ & $0.71 \pm 0.01$  \\ 
Dynamic & $0.63 \pm 0.01$ & $0.70 \pm 0.01$ & $0.72 \pm 0.01$ & $0.76 \pm 0.00$ \\ 
Dynamic (adjusted) & $0.67 \pm 0.01$ & $0.71 \pm 0.01$ & $0.72 \pm 0.01$ & $0.76 \pm 0.00$ \\
\bottomrule 
\end{tabular}
\end{table*}

\answerVRone{

\subsection{Adjusting to Different Hardware}
\label{sec:hardwareajustmentexperiment}
To evaluate the described calibration approach in Section~\ref{sec:automl_features} that allows us to apply \system{} on any machine, we conduct additional experiments on a powerful computer with Intel(R) Core(TM) i7-8565U CPU @ 1.80 GHz and 38 GB RAM. To benchmark both machines, we run \system{} Static for the dataset ``riccardo'' for 10min for 10 times and measure the average validation balanced accuracy across the search time as reported in Figure~\ref{figure:bench}. We choose this dataset because it has 20k instances and 4k features, and it takes more time to converge to find a well-performing model. If the data is too simple, we could not compare the convergence across time well.
Then, we conduct the experiments for search times from 10s to 5min on the new machine with and without hardware adjustment as reported in Table~\ref{tab:benchres}.
First, we see that even without hardware adjustment the \system{} Dynamic still outperforms the static one. 
With hardware adjustment, for 10s and 30s, the average balanced accuracy improves by 4\% and 1\% accordingly.
So, we conclude for small search budgets, hardware adjustment does improve our system. This finding also reduces the cost of the offline benchmark because \system{} can run the benchmark for at most 10 minutes and turn off mapping for larger search times.
}

\begin{figure}
	\centering
	\input{eval_constraints}
	\caption{\answerVRone{Meta-Training Model F1 score under a varying number of constraints and different durations of meta-training.}}
	\label{figure:scalingconstraints}
\end{figure}
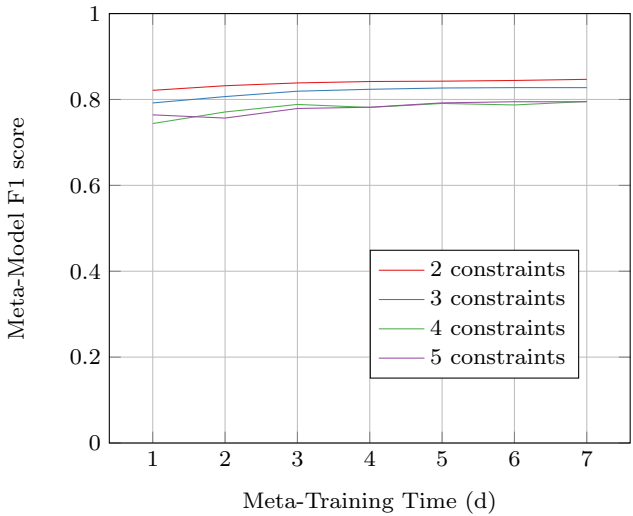

\answerVRone{

\subsection{Impact of the Number of Constraints on Meta-Training}
\label{sec:numberconstraintsmetatraining}
To analyze the impact of the number of constraints on the meta-learning performance, we run meta-training with up to 5 constraints, 4 ML application constraints 
and the mandatory search time constraint.

To compare the performance across constraint combinations, for each combination, we randomly pick random configurations, constraint thresholds, and datasets and evaluate whether the corresponding configuration performs better than the default configuration. We gather this test set for 1 week for each constraint combination.
Then, for each number of constraints, we apply our alternating meta-training for 1 week. Finally, we report the F1 score of the meta-model on each test set reporting whether the corresponding configuration outperformed the default configuration at least once. We report the corresponding averaged F1 scores across 5 repetitions for each day of the week of meta-training in Figure~\ref{figure:scalingconstraints}.

With increasing meta-training duration, the F1 score of the meta-model increases for all considered numbers of constraints. After one week, all constraint combinations reach an F1 score higher than $79\%$.
As expected, by adding additional constraints, the F1 score slightly decreases by up to 3\%, which is rather marginal considering the multiplied complexity of the resulting search space.  
This analysis can help to estimate how much time it may take to generate sufficient training data for larger search spaces.
}

%% file: 10s.tex
\begin{minipage}{.3\textwidth}
{\tiny
\begin{forest} for tree={
    grow=east
}
[
    [augmentation
        [Random]
        [ADASYN]
    ]
    [clf		
        [RF
            [min samples split]
            [bootstrap]
        ]
        [AdaB.
            [algorithm: SAMME]
        ]
        [B.NB
            [alpha]
        ]
        [DT
            [criterion: entropy]
            [max depth factor]
        ]
        [E.Trees
            [criterion]
            [min samples leaf]
            [bootstrap: False]
        ]
        [G.NB]
        [LSVC
            [loss: hinge]
        ]
        [M.NB
            [fit prior: False]
        ]
        [SVC
            [kernel
                [poly]
                [sigmoid]
            ]
            [gamma]
            [coef0]
            [shrinking
                [True]
                [False]
            ]
            [tol]
        ]       
    ]
    [cat. enc.
        [One-Hot]
        [Frequency]
    ]
]
\end{forest}}
\end{minipage}

%% file: 1min.tex
\begin{minipage}{.3\textwidth}
{\tiny
\begin{forest} for tree={
    grow=east
}
[
    [augmentation
        [Random]
    ]
    [clf		
        [AdaB.
            [n estimators]
            [algorithm
                [SAMME.R]
                [SAMME]
            ]
        ]
        [DT
            [criterion]
            [min samples split]
            [min samples leaf]
        ]
        [E.Trees
            [max features]
            [min samples split]
            [min samples leaf]
            [bootstrap: False]
        ]
        [G.NB]
        [M.NB]
        [QDA
            [reg param]
        ]
        [SGD
            [alpha]
            [l1 ratio]
            [tol]
            [learning rate]
        ]
        [SVC
            [kernel
                [rbf]
                [sigmoid]
            ]
            [tol]
        ]
        [LDA
            [shrinkage factor]
        ]
    ]
    [scaler
        [Normalizer]
        [QuantileTransformer]
        [RobustScaler
            [q max]
        ]
    ]
    [cat. enc.: Label]
]
\end{forest}}
\end{minipage}

%% file: 5min.tex
\begin{minipage}{.3\textwidth}
{\tiny
\begin{forest} for tree={
    grow=east
}
[
    [clf		
        [AdaB.
            [n estimators]
            [learning rate]
            [algorithm: SAMME]
            [max depth]
        ]
        [B.NB]
        [DT
            [criterion]
            [max depth factor]
            [min samples split]
        ]
        [E.Trees
            [criterion]
            [min samples split]
            [min samples leaf]
            [bootstrap: True]
        ]
        [G.NB]
        [KNN
            [p]
        ]
        [M.NB
            [fit prior
                [True]
                [False]
            ]
        ]
        [PA
            [C]
        ]
        [SGD
            [l1 ratio]
            [epsilon]
            [eta0]
            [average: False]
        ]
        [SVC
            [coef0]
            [tol]
        ]           
    ]
    [cat. enc.: One-Hot]
]
\end{forest}
}
\end{minipage}

%% file: eval_constraints.tex
\begin{tikzpicture}[scale=1.0]
\begin{axis}[
  ymin=0, ymax=1,
        grid=both,
        xlabel=Meta-Training Time (d),
        ylabel=Meta-Model F1 score,
        legend entries={
	2 constraints, 3 constraints, 4 constraints, 5 constraints},
        legend style={at={(0.5,0.3)},anchor=west}]
\addplot[color=r1] coordinates {(1,0.8212800322313896)(2,0.8320073895810574)(3,0.8385086082973461)(4,0.8418242005971805)(5,0.8425853935440453)(6,0.8442502003880087)(7,0.846866832249033)};
\addplot[color=r2] coordinates {(1,0.7919454484075097)(2,0.8066996721375226)(3,0.8192364585594769)(4,0.8236947262731202)(5,0.8266901730881286)(6,0.8276169030861759)(7,0.8275555878763508)};
\addplot[color=r3] coordinates {(1,0.7439027984311791)(2,0.7709083763392481)(3,0.7884540660935945)(4,0.7814632517941087)(5,0.7907539273863986)(6,0.7872202628256652)(7,0.795382340275223)};
\addplot[color=r4] coordinates {(1,0.7641860008531018)(2,0.7566762062455912)(3,0.7789526048517731)(4,0.7819707168935868)(5,0.7920501888821159)(6,0.7946914203659502)(7,0.7945840922219833)};
\end{axis}
\end{tikzpicture}

%% file: 05_related_work.tex
\section{Related Work}
\label{sec:relatedwork}
Our work on constraint-driven AutoML combines research from various areas of optimization, AutoML, and meta-learning.

\noindent\textbf{Constrained Optimization.} One direction of work addresses constrained optimization by learning a surrogate model that estimates whether sampled configurations violate the corresponding constraints~\cite{spearmint,perrone2020fair,gpflowopt,DBLP:journals/jmlr/AriafarCBD19,DBLP:conf/aaai/0001RV0BSW0G20}. However, this approach has two downsides. First, it requires the surrogate models to learn the constraints each time from scratch. Second, it cannot adjust the parameters of the AutoML systems, such as the validation approach or the search strategy, to the corresponding ML task.

\noindent\textbf{Meta-Learning.} 
A more effective approach is to learn upfront whether a given ML pipeline satisfies a well-known constraint, such as training time~\cite{mohr2021predicting}. This approach does not require learning the constraint each time from scratch. Still, it does not adjust the AutoML parameters.
Another direction is to meta-optimize the AutoML parameters. For instance, Lindauer et al.~\cite{DBLP:journals/corr/abs-1908-06674} optimize the parameters of hyperparameter optimization. However, they do not consider constraints. Further, Auto-Sklearn~2~\cite{AutoSklearn2} only supports predicting discrete strategy decisions using pair-wise modeling. Therefore, their approach does not support continuous AutoML hyperparameters and does not scale to hundreds of settings. This scalability issue also hinders joint strategy prediction as the combinatorial space is too huge.
Van Rijn et al. leverage meta-learning to identify the most important hyperparameter for various ML models individually~\cite{DBLP:conf/kdd/RijnH18}. However, they do not consider constraints.
Alpine Meadow~\cite{shang2019democratizing} uses the history of the quality and cost of all so far run pipelines to warm-start search, but can also not handle constraints.

\noindent\textbf{Accelerating AutoML.} Further, there is a large effort in the data management community to speed up AutoML systems. For instance, Li et al. propose to leverage search space decomposition~\cite{DBLP:journals/pvldb/LiSZJLDZY00021}. Yakovlev et al. propose to leverage proxy models, iteration-free optimization, and adaptive data reduction to accelerate hyperparameter optimization~\cite{yakovlev2020oracle}. Another well-known approach to speed up hyperparameter optimization is to leverage successive halving~\cite{DBLP:journals/jmlr/LiJDRT17,DBLP:conf/icml/FalknerKH18}. It starts by evaluating many configurations on a small budget and incrementally chooses the best half of the configurations to evaluate them on a bigger budget. Xin et al. leverage caching to accelerate hyperparameter optimization~\cite{xin2018helix}. However, their strategies cannot be applied in case of validation split reshuffling. Nakandala et al. propose a new parallel SGD execution strategy to speed up hyperparameter optimization for SGD-based models~\cite{DBLP:journals/pvldb/NakandalaZK20}.
\answerVRthree{Hilprecht et al.~\cite{DBLP:conf/deem/HilprechtHR0B23} propose to make ML pipelines end-to-end differentiable to avoid costly Bayesian optimization.
SystemDS~\cite{boehm2019systemds,DBLP:journals/pvldb/BoehmDEEMPRRSST16} allows users to specify ML programs in a declarative R-like language and compiles it to highly efficient hardware-specific code that can be distributed.
Shah et al.~\cite{DBLP:conf/sigmod/ShahLKY021} extensively benchmark feature type detection that is important because the downstream AutoML system is dependent on the right feature type classification.
}
The aforementioned systems and algorithms are orthogonal to our contribution as they do not consider the search space of AutoML but optimize the computation for training and parameter tuning. 


%% file: 06_conclusion.tex
\section{Conclusion}
We proposed integrating constraints as a first-class citizen into AutoML - a paradigm that we call constraint-driven AutoML. As the constraints set limitations on the hyperparameter search, we proposed an approach to dynamically change the AutoML search space for the constraints at hand.
To achieve this goal, we leverage active meta-learning.
To explore the huge space of datasets, AutoML configurations, and constraints, we sample those combinations that benefit the meta-model.
To show the full benefit of this approach, we develop a simple adjustable AutoML system, \system{}, that exposes its whole ML hyperparameter space as binary AutoML parameters to have a task-specific search space. 
This way, \emph{\system{} Dynamic} can decide for every single ML hyperparameter whether it should be optimized or not.
It automatically chooses an ML hyperparameter space for search time constraints that is similar to the space covered by the hand-tuned Auto-Sklearn2~\cite{AutoSklearn2} system. Overall, our new approach allows for configurable generic AutoML systems that dynamically adjust to the task and constraints at hand, and thus further increase the applicability of AutoML systems in practical application.